\documentclass[10pt,twocolumn,letterpaper]{article}

\usepackage[pagenumbers]{cvpr} 

\usepackage{graphicx}
\usepackage{amsmath}
\usepackage{amssymb}
\usepackage{booktabs}
\usepackage{xcolor}
\usepackage{xspace}
\usepackage{enumerate}

\usepackage{multirow}
\usepackage{makecell}
\usepackage{colortbl}

\newcolumntype{x}[1]{>{\centering\arraybackslash}p{#1pt}}
\newcolumntype{y}[1]{>{\raggedright\arraybackslash}p{#1pt}}
\newcolumntype{z}[1]{>{\raggedleft\arraybackslash}p{#1pt}}

\usepackage[pagebackref,breaklinks,colorlinks]{hyperref}

\usepackage[capitalize]{cleveref}
\crefname{section}{Sec.}{Secs.}
\Crefname{section}{Section}{Sections}
\Crefname{table}{Table}{Tables}
\crefname{table}{Tab.}{Tabs.}

\def\cvprPaperID{305} 
\def\confName{CVPR}
\def\confYear{2023}
\newcommand{\name}{ViRReq\xspace}

\begin{document}

\title{Visual Recognition by Request}

\author{%
  Chufeng Tang$^1$ \quad Lingxi Xie$^2$ \quad Xiaopeng Zhang$^2$ \quad Xiaolin Hu$^1$ \quad Qi Tian$^2$ \\
  $^1$Dept. of Comp. Sci. \& Tech., State Key Lab for Intell. Tech. \& Sys., \\ Institute for AI, BNRist, Tsinghua University \qquad
  $^2$Huawei Inc.\\
}

\maketitle

\begin{abstract}
Humans have the ability of recognizing visual semantics in an unlimited granularity, but existing visual recognition algorithms cannot achieve this goal. In this paper, we establish a new paradigm named \text{visual recognition by request} (\text{\name}\footnote{We recommend the readers to pronounce \name as \textsf{/'virik/}.}) to bridge the gap. The key lies in decomposing visual recognition into atomic tasks named \text{requests} and leveraging a \text{knowledge base}, a hierarchical and text-based dictionary, to assist task definition. \name allows for (i) learning complicated whole-part hierarchies from highly incomplete annotations and (ii) inserting new concepts with minimal efforts. We also establish a solid baseline by integrating language-driven recognition into recent semantic and instance segmentation methods, and demonstrate its flexible recognition ability on CPP and ADE20K, two datasets with hierarchical whole-part annotations.
\end{abstract}

\section{Introduction}
\label{introduction}

Visual recognition is one of the fundamental problems in computer vision. In the past decade, visual recognition algorithms have been largely advanced with the availability of large-scale datasets and deep neural networks~\cite{krizhevsky2012imagenet,he2016deep,dosovitskiy2021image}. Typical examples include the ability of recognizing $10\rm{,}000$s of object classes~\cite{deng2009imagenet}, segmenting objects into parts or even parts of parts~\cite{zhou2019semantic}, using natural language to refer to open-world semantic concepts~\cite{radford2021learning}, \textit{etc}.

\begin{figure}
\centering
\includegraphics[width=8cm]{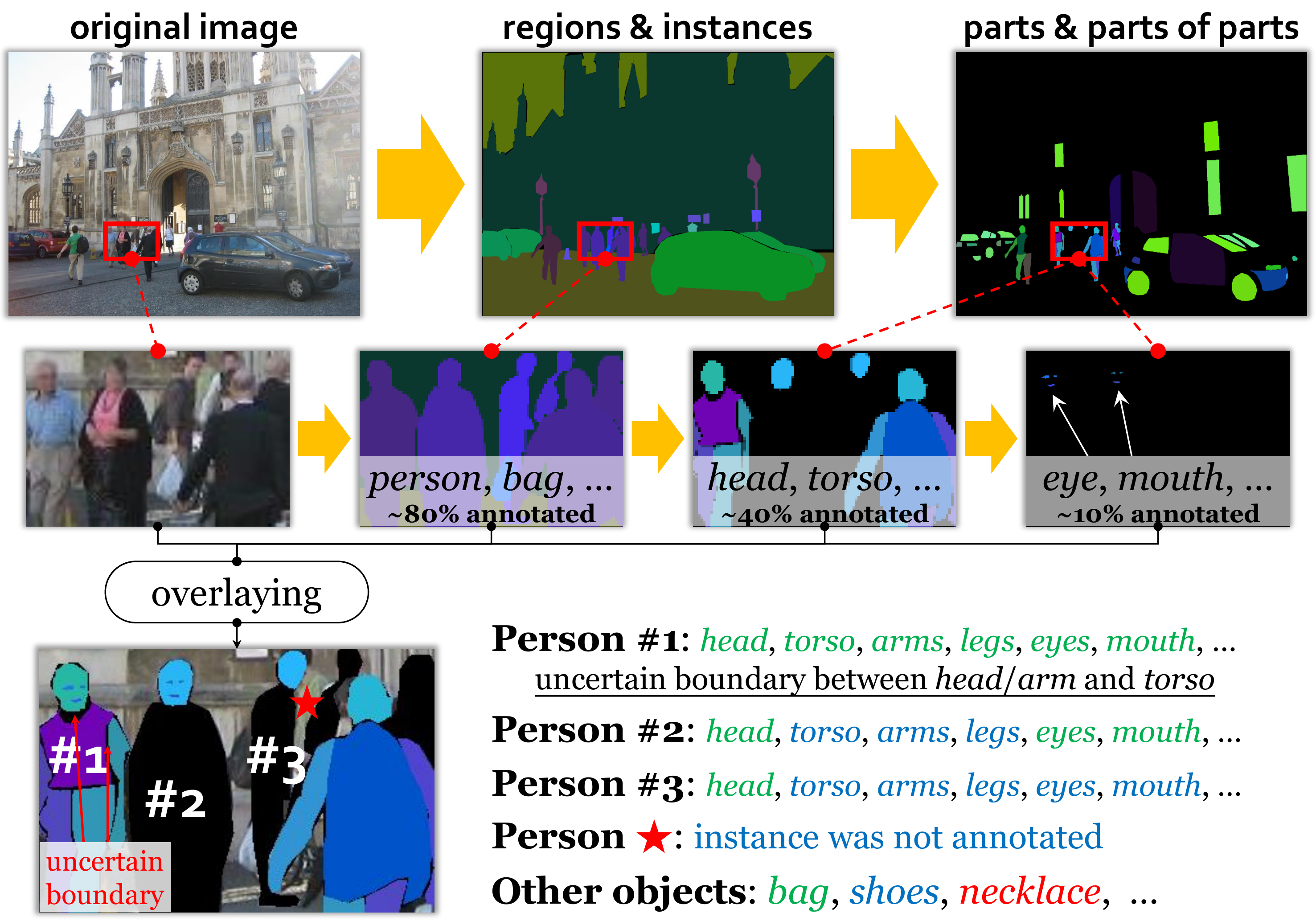}
\vspace{-0.2cm}
\caption{An illustration of unlimited granularity in visual recognition. \textbf{Top}: an example image from ADE20K~\cite{zhou2019semantic} with instance-level and part-level annotations. \textbf{Middle}: more and more incomplete annotations of instances, parts, and parts of parts. \textbf{Bottom}: as granularity goes finer, higher uncertainty occurs in recognizing the boundary (left) and semantic class (right) of objects and/or parts -- here, \textcolor{green}{green}, \textcolor{blue}{blue} and \textcolor{red}{red} texts indicate \textcolor{green}{labeled}, \textcolor{blue}{unlabeled (but defined)}, and \textcolor{red}{unlabeled (and undefined)} objects, respectively.}
\label{fig:granularity}
\end{figure}

Despite the increasing recognition accuracy in standard benchmarks, we raise a critical yet mostly uncovered issue, namely, the \textbf{unlimited granularity} of visual recognition. As shown in Figure~\ref{fig:granularity}, given an image, humans have the ability of recognizing arbitrarily fine-grained contents from it, but existing visual recognition algorithms cannot achieve the same goal. Superficially, this is caused by limited annotation budgets so that few training data is available for fine-grained and/or long-tailed concepts, but we point out that a more essential reason lies in the conflict between granularity and recognition certainty -- as shown in Figure~\ref{fig:granularity}, \textit{when annotation granularity goes finer, annotation certainty will inevitably be lower}. This motivates us that the granularity of visual recognition shall be variable across instances and scenarios. For this purpose, we propose to interpret semantic annotations into smaller units (\textit{i.e.}, requests) and assume that recognition is performed only when it is asked, so that one can freely adjust the granularity according to object size, importance, clearness, \textit{etc}.

In this paper, we establish a new paradigm named \textbf{visual recognition by request} (\textbf{\name}). We only consider the segmentation task in this paper because it best fits the need of unlimited granularity in the spatial domain. Compared to the conventional setting, the core idea is to decompose visual recognition into a set of atomic tasks named \textbf{requests}, each of which performs one step of recognition. Specifically, there are two request types, with the first type decomposing an instance into semantic parts and the second type segmenting an instance out from a semantic region. For the first type, a \textbf{knowledge base} is available as a hierarchical text-based dictionary to guide the segmentation procedure.

Compared to the existing paradigms of visual recognition, \name has two clear advantages. First, \name naturally has the ability of learning complicated visual concepts (\textit{e.g.}, the whole-part hierarchy in ADE20K~\cite{zhou2019semantic}) from highly incomplete annotations (\textit{e.g.}, part-level annotations are available for only a small amount of instances), while conventional methods can encounter several difficulties. Second, \name allows a new visual concept (\textit{e.g.}, objects, parts, \textit{etc.}) to be added by simply updating the knowledge base and annotating a few training examples. We emphasize that the change of knowledge base does not impact the use of existing training data as each sample is bound to a specific version of the knowledge base. This property, called data versioning, allows for incremental learning with all historical data available.

To deal with \name, we build a query-based recognition algorithm that (i) extracts visual features from the image, (ii) computes text embedding vectors from the request and knowledge base, and (iii) performs interaction between image and text features. The framework is highly modular and the main parts (\textit{e.g.}, feature extractors) can be freely replaced. We evaluate \name with the proposed recognition algorithm on two datasets, namely, the CPP dataset~\cite{degeus2021part} that extends Cityscapes~\cite{cordts2016cityscapes} with part-level annotations and the ADE20K dataset~\cite{zhou2019semantic} that offers a multi-level hierarchy of complicated visual concepts. We parse the annotations of each image into a set of requests for training, and define a new evaluation metric named hierarchical panoptic quality (HPQ) for measuring the segmentation accuracy. Thanks to the ability of learning from incomplete annotations, \name can report part-aware segmentation accuracy on ADE20K, which, to the best of our knowledge, is the first ever work to achieve the goal. In addition, \name shows a promising ability of open-domain recognition, including absorbing new visual concepts (\textit{e.g.}, objects, parts, \textit{etc.}) from a few training samples and understanding anomalous or compositional concepts without training data.

In summary, the contribution of this paper is three fold: (i) pointing out the issue of unlimited granularity, (ii) establishing the paradigm named visual recognition by request, and (iii) setting up a solid baseline for this new direction.

\section{Preliminaries and Insights}
\label{preliminaries}

\subsection{Related Work in Visual Recognition}
\label{preliminaries:relatedwork}

In the deep learning era~\cite{lecun2015deep}, with deep neural networks~\cite{krizhevsky2012imagenet,he2016deep,dosovitskiy2021image,liu2021swin} being adopted as a generalized tool for representation learning, the community has been pursuing for an effective method to improve the \textit{granularity} of visual recognition, which we refer to the ability of recognizing rich visual contents. From the perspective of defining a fine-grained visual recognition task, typical examples include collecting image data for more semantic classes (\textit{e.g.}, ImageNet~\cite{deng2009imagenet} has more than $22\mathrm{K}$ classes) and labeling finer parts (\textit{e.g.}, ADE20K~\cite{zhou2019semantic} labeled more than $600$
classes of parts and parts of parts). Despite the great efforts, these high-quality datasets are still far from the goal of \textit{unlimited granularity} of visual recognition, in particular, recognizing everything that humans can recognize~\cite{xie2021considered}. In what follows, we categorize existing recognition approaches into two types and analyze their drawbacks \textit{assuming the goal being unlimited granularity}.

The first type is the \textbf{classification-based} tasks which refer to the targets of assigning a class ID for each semantic unit. The definition of semantic units determines the visual recognition task, \textit{e.g.}, the unit is an image for image classification~\cite{deng2009imagenet}, a rectangular region for object detection~\cite{everingham2010pascal,lin2014microsoft,shao2019objects365}, a masked region for semantic/instance segmentation~\cite{everingham2010pascal,lin2014microsoft,cordts2016cityscapes,zhou2019semantic}, a keypoint for human pose estimation~\cite{andriluka20142d,lin2014microsoft}, \textit{etc}. This mechanism has a critical drawback, \textit{i.e.}, the conflict between granularity and certainty. As shown in Figure~\ref{fig:granularity}, the certainty (also accuracy) of annotation is not guaranteed when either semantic or spatial granularity goes finer. To alleviate the conflict, we shall allow the granularity to vary across semantic units, \textit{e.g.}, large objects are labeled with parts but small objects are not. \textit{In this paper, we propose a framework that annotations are made upon request.}

The second type is the \textbf{language-driven} tasks which leverage texts or other modalities to refer to specific visual contents. Typical examples include visual question answering~\cite{antol2015vqa,wu2017visual,das2018embodied,gordon2018iqa}, image captioning~\cite{vinyals2015show,hossain2019comprehensive}, visual grounding~\cite{hu2016natural,mao2016generation,zhu2016visual7w,liu2017recurrent}, visual reasoning~\cite{zellers2019recognition}, \textit{etc}. Recently, with the availability of vision-language pre-trained models that learn knowledge from image-text pairs (\textit{e.g.}, CLIP~\cite{radford2021learning}, GLIP~\cite{li2022grounded}, \textit{etc.}), this mechanism has shown promising abilities for open-world recognition~\cite{joseph2021towards,zareian2021open,minderer2022simple}, especially in using language as queries for visual recognition~\cite{zhou2022learning,li2022language,ding2022open}. However, the ability of natural language of referring to detailed visual contents (\textit{e.g.}, the parts of a specific \textsf{person} in a complex scene with tens of \textsf{persons}) is very much limited, hence, it is unlikely that purely relying on language can achieve unlimited granularity. \textit{In this paper, we use language to define a hierarchical dictionary but establish the benchmark mainly based on vision itself.}

\subsection{Key Insights towards Unlimited Granularity}
\label{preliminaries:insights}

Based on the above analysis, we have learned that unlimited granularity is not yet achieved by existing paradigms. We summarize two key insights towards the goal, based on which we present our solution in the next section.

First, we note that unlimited granularity includes the scope of openness (\textit{i.e.}, the open-domain property) and is more challenging. Currently, leveraging data from other modalities (\textit{e.g.}, natural language) is the most promising and convenient way of introducing openness, so we use texts as labels to define semantic concepts. This strategy makes it easier to capture the relationship between concepts (\textit{e.g.}, different kinds of \textsf{vehicles} such as \textsf{car} and \textsf{bus} are closely related) and to learn compositional concepts (\textit{e.g.}, transferring the visual features from the parts of \textsf{car} to the parts of \textsf{bus}).


Second and more importantly, we assume that the recognition granularity is variable across instances and/or scenarios. This calls for a flexible recognition paradigm that \textit{not always} pursues for the finest granularity but performs recognition only with proper requests (\textit{e.g.}, a \textsf{person} can be annotated with parts or even parts of parts if the resolution is sufficiently large), yet the instance can be ignored if the resolution is very small. \textit{In our proposal, the granularity is freely controlled by decomposing the recognition task into requests.}

\section{Problem: Visual Recognition by Request}
\label{problem}

As shown in Figure~\ref{fig:setting}, \name offers a novel and unified paradigm for both data annotation and visual recognition. Throughout this paper, we parse existing datasets (CPP and ADE20K) into the \name setting and present a solid baseline. We advocate that the community annotates and organizes visual data using this paradigm, so as to push visual recognition towards unlimited granularity.


\subsection{Notations and Task Definition}
\label{problem:definition}

We consider the image segmentation task in this paper because it mostly fits our goal towards unlimited granularity. Conventional segmentation benchmarks often provide a pre-defined dictionary that contains all visual concepts (\textit{i.e.}, classes). Given an image, $\mathbf{X}=\left\{x_{w,h}\right\}_{w=1,h=1}^{W,H}$, where $x_{w,h}$ indicates the pixel at the position $\left(w,h\right)$ and $W,H$ are the image width and height, respectively, it is required to predict a pixel-wise segmentation map, $\mathbf{Z}=\left\{z_{w,h}\right\}_{w=1,h=1}^{W,H}$, where $z_{w,h}$ represents the semantic and/or instance labels of $x_{w,h}$. As we have analyzed
above, such a definition can encounter difficulties when the scene becomes complex and the granularity becomes finer.

The core idea of \name is to decompose the recognition task, $\mathbf{X}\mapsto\mathbf{Z}$, into a series of requests, denoted as $\mathcal{R}=\left\{\mathbf{r}_k\right\}_{k=1}^K\doteq\left\{Q_k,\mathcal{A}_k\right\}_{k=1}^K$, where $Q_k$ and $\mathcal{A}_k$ denote the $k$-th request and answer, respectively. These requests are performed sequentially, and some of them may depend on the recognition results of earlier requests, \textit{e.g.}, the system must first segment an instance before further partitioning it into parts. The segmentation results are stored in a tree, denoted as $\mathcal{T}=\left\{\mathbf{t}_l\right\}_{l=0}^L\doteq\left\{\mathbf{M}_l,c_l,\tau_l,\mathcal{U}_l\right\}_{l=0}^L$, where each node is a semantic region or an instance. For the $l$-th node, $\mathbf{M}_l\in\left\{0,1\right\}^{W\times H}$ is the binary mask, $c_l\in\left\{1,2,\ldots,C\right\}$ is the class index in the knowledge base (see the next paragraph), $\tau_l\in\left\{0,1\right\}$ indicates whether the node corresponds to an instance, and $\mathcal{U}_l\subset\left\{l+1,\ldots,L\right\}$ is the set of child nodes of $\mathbf{t}_l$. If $\mathbf{t}_{1}$ is the parent node of $\mathbf{t}_{2}$, then $\mathbf{t}_{2}$ is recognized by answering a request on $\mathbf{t}_{1}$. In our definition, an instance must be a child node of a semantic region. Note that $\mathcal{R}$ and $\mathcal{T}$ are tightly related: each non-leaf node $\mathbf{t}_l$ in $\mathcal{T}$ corresponds to a request in $\mathcal{R}$, say $\mathbf{r}_k$, and the answer $\mathcal{A}_k$ corresponds to the set of all child nodes of $\mathbf{t}_l$.

\begin{figure*}
\centering
\includegraphics[width=16cm]{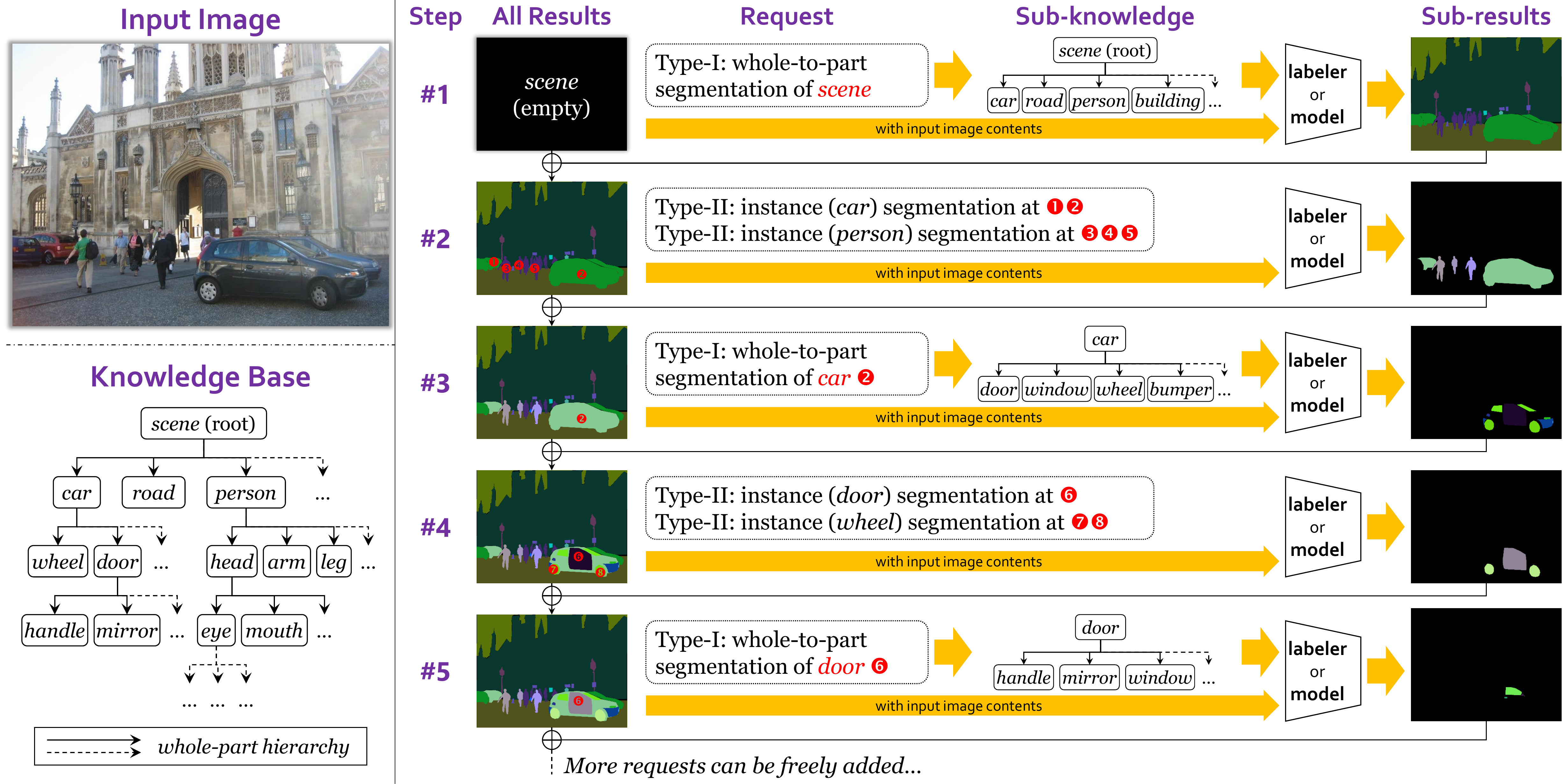}
\vspace{-0.3cm}
\caption{An illustration of the overall pipeline of \name that is built upon a pre-defined knowledge base. Given an input image, \name starts with an empty field (\textit{i.e.}, nothing is annotated or recognized) and iteratively executes Type-I/II requests towards recognition results of finer granularity. The sub-knowledge for Type-I requests is extracted from the knowledge base. The labeler can stop after any number of steps, yet the model can learn from incomplete annotations without difficulties. To save space, we combine a few Type-II requests into one step. We only display five steps here, while a typical annotation/recognition procedure often include much more steps (and requests).}
\label{fig:setting}
\end{figure*}

Based on the above definition, a typical procedure of visual recognition is described as follows and illustrated in Figure~\ref{fig:setting}. Initially, nothing is annotated and $\mathcal{T}$ has only a root node $\mathbf{t}_0$, in which $\mathbf{M}_0=\mathbf{1}^{W\times H}$, $c_0=1$ (indicating a special class named \textsf{scene}), $\tau_0=1$ (\textit{i.e.}, this is an instance of \textsf{scene}), and $\mathcal{U}_0=\varnothing$. Each request, $\mathbf{r}_k$, finds the corresponding node, $\mathbf{t}_l$, performs the recognition task, translates the answer into child nodes of $\mathbf{t}_l$, and adds them back to $\mathcal{T}$. There are two types of requests:
\begin{enumerate}[Type-I]
\item \textbf{Whole-to-part semantic segmentation}. Prerequisite: $\mathbf{t}_l$ must be an instance (\textit{i.e.}, $\tau_l=1$). The system fetches the class label, $c_l$, and looks up the dictionary for the set of its sub-classes in texts (\textit{e.g.}, \textsf{person} has sub-classes (parts) of \textsf{head}, \textsf{torso}, \textsf{arm}, and \textsf{leg}, \textit{etc.}).
\item \textbf{Instance segmentation}. Prerequisite: $\mathbf{t}_l$ must be a semantic region (\textit{i.e.}, $\tau_l=0$). Each time, a probing pixel (named probe) $\left(a_k,b_k\right)$ is provided with the request, $M_l\!\left(a_k,b_k\right)=1$, and the task is to segment the instance that occupies this probe. Note that, in the testing stage, we can fairly compare with conventional instance segmentation methods using densely sampled probes -- see Section~\ref{experiments:cpp}.
\end{enumerate}

At the beginning, the entire image is regarded as a special instance named \textsf{scene}, and the top-level semantic classes (\textit{e.g.}, \textsf{sky}, \textsf{building}, \textsf{person}, \textit{etc.}) are considered as its parts. We use such definition to simplify the overall notation system. Briefly, Type-I requests segment instances into semantics, while Type-II requests find instances from semantics. By executing them iteratively and alternately (\textit{i.e.}, the path that yields one unit must be Type-I$\rightarrow$Type-II$\rightarrow$Type-I$\rightarrow$...), one can segment arbitrarily fine-grained units (\textit{i.e.}, towards unlimited granularity) from the input image.

To define the whole-part hierarchy for Type-I requests, we establish a directed graph known as the \textbf{knowledge base} -- a toy example is shown in Figure~\ref{fig:setting}. Each node of the graph is an object/part concept, and it may have a few child nodes that correspond to its named parts. In our formulation, the class labels appear as texts rather than a fixed integer ID -- this is to ease the generalization towards open-domain recognition: given a pre-trained text embedding (\textit{e.g.}, CLIP~\cite{radford2021learning}), some classes are recognizable by language even if they never appear in the training set. The root class is \textsf{scene} that corresponds to the entire picture.

\subsection{Advantages over Existing Paradigms}
\label{problem:advantages}

\name is different from existing visual recognition paradigms, in particular, language-driven visual recognition mentioned above. We decompose visual recognition into requests, guided by the knowledge base, to pursue for the ultimate goal, \textit{i.e.}, visual recognition of unlimited granularity. This brings the following two concrete advantages.

First, from a \textbf{micro} view, \name allows us to learn complex whole-part hierarchies from highly incomplete annotations. As shown in Figure~\ref{fig:setting}, the parsed requests do not deliver noisy supervision even though the training data (i) ignores many instances for dense objects and/or (ii) annotates fine-grained parts only for a small amount of objects. More importantly, it alleviates the conflict between granularity and certainty by only annotating the contents that labelers are sure about. In other words, \name boosts certainty by sacrificing completeness for a single image, but the entire dataset covers the knowledge base sufficiently well. 

Second, from a \textbf{macro} view, \name allows us to add new concepts (objects and/or parts) easily. To do this, one only needs to (i) insert a text-based node to the knowledge base and (ii) annotate a few images that contains the concept. Note that although the knowledge base augments with time, this does not prevent us from using old training data, because we can associate each training image to the knowledge base that was used for annotating new concepts (which we call data versioning). That said, the new concept can remain unlabeled in old images even it appears, as long as the old images are associated to the old knowledge base that does not contain the new concept. In Section~\ref{experiments:ade}, we will show the benefits of data versioning in the incremental learning experiments in ADE20K.

\section{Query-Based Recognition: A Baseline}
\label{approach}

\name calls for an algorithm that deals with requests $\left\{\mathbf{r}_k\right\}_{k=1}^K$ one by one, similar to the illustration in Figure~\ref{fig:setting}. Specifically, in each step, the input involves the image $\mathbf{X}$, the current recognition result $\mathbf{Z}_k$ (prior to the $k$-th request), the request $Q_k$, and the knowledge base. Processing each request involves extracting visual features, constructing queries, performing recognition and filtering, and updating the current recognition results.

\noindent
\textbf{Visual features.}\quad
We extract visual features from $\mathbf{X}$ using a deep neural network (\textit{e.g.}, a conventional convolutional neural network~\cite{krizhevsky2012imagenet,he2016deep} or vision transformer~\cite{dosovitskiy2021image,liu2021swin}), obtaining a feature map $\mathbf{F}\in\mathbb{R}^{H'\times W'\times D}$, where $H'\times W'$ is usually a down-sampled scale of the original image.

\noindent
\textbf{Language-based queries.}\quad
Each request is transferred into a set of query embedding vectors $\mathbf{E}$, for which a pre-trained text encoder (\textit{e.g.}, CLIP~\cite{radford2021learning}) and the knowledge base are required. \textbf{(1)} For Type-I requests (\textit{i.e.}, whole-to-part semantic segmentation), the target class (in text, \textit{e.g.}, \textsf{person}) is used to look up the knowledge base, and a total of $P_k$ child nodes are found (in text, \textit{e.g.}, \textsf{head}, \textsf{arm}, \textsf{leg}, \textsf{torso}). We feed them into the text encoder to obtain $P_k$ embedding vectors, $\mathbf{E}=\left[\mathbf{e}_{k,1};\ldots;\mathbf{e}_{k,P_k}\right]$. \textbf{(2)} For Type-II requests (\textit{i.e.}, instance segmentation), given a probing pixel, a triplet $\{a_k,b_k,c_k\}$ is obtained, where $(a_k,b_k)$ is the pixel coordinates and $c_k$ is the target class that is determined by the current recognition result. To construct the query, the target class (in text) is directly fed into the text encoder to obtain the semantic embedding vector  $\mathbf{e}_k$, and the pixel coordinates are fed into a positional encoder to obtain the positional embedding $\mathbf{p}_k$, where $\mathbf{p}_k=(a_k,b_k)$ for simplicity in our implementation. The query embedding is obtained by combining them, $\mathbf{E}=(\mathbf{e}_k, \mathbf{p}_k)$. All text embedding vectors are $D$-dimensional, \textit{i.e.}, same as the visual features.

\noindent
\textbf{Language-driven recognition.}\quad
Visual features $\mathbf{F}$ are interacted with the language-based queries $\mathbf{E}$ to perform segmentation. \textbf{(1)} For Type-I requests, we directly compute the inner-product between the visual feature vector $\mathbf{f}_{w,h} \in \mathbb{R}^{D}$ of each pixel and the embedding vectors $\mathbf{E}$, obtaining a $P_k$-dimensional class score vector, $\mathbf{u}_{w,h}=\mathbf{E}^\top\cdot\mathbf{f}_{w,h}$. The entry with the maximal response is the predicted class label. To allow open-set recognition, we augment $\mathbf{u}_{w,h}$ with an additional entry of $0$, \textit{i.e.}, $\hat{\mathbf{u}}_{w,h}={\left[\mathbf{u}_{w,h};0\right]}$, where the added entry stands for the \textsf{others} class -- that said, if the responses of all $P_k$ normal entries are smaller than $0$, the corresponding position is considered an unseen (anomalous) class -- see such examples in Section~\ref{experiments:ade}. \textbf{(2)} For Type-II requests, different from semantic segmentation, existing instance segmentation methods usually generate massive proposals (\textit{e.g.}, box proposals in Mask R-CNN~\cite{he2017mask}, feature locations in CondInst~\cite{tian2020conditional}) and predict the class label, bounding box, and binary mask for each proposal. To achieve instance segmentation \textit{by request}, we first select the spatially related proposals by the positional embedding $\mathbf{p}_k$. Specifically, feature locations (in CondInst) near the probing pixel at each feature pyramid level are selected for subsequent prediction. Prediction with the highest categorical score, obtained as similar as processing Type-I requests (inner-product with text embedding), is chosen as the final result.

\noindent
\textbf{Top-down filtering.}\quad
We combine the request $r_k$ and the current recognition result $\mathbf{Z}_k$ to constrain the segmentation masks predicted above. That said, we follow the top-down criterion to deal with the segmentation conflict of different levels requests. For example, part segmentation results should strictly inside the corresponding instance region. Applying advanced mask fusion methods~\cite{li2018learning,li2019attention,mohan2021efficientps,ren2021refine} may produce better results. The filtered masks are added back to the current recognition results, which lays the foundation of subsequent requests.

\begin{figure}
\centering
\includegraphics[width=\linewidth]{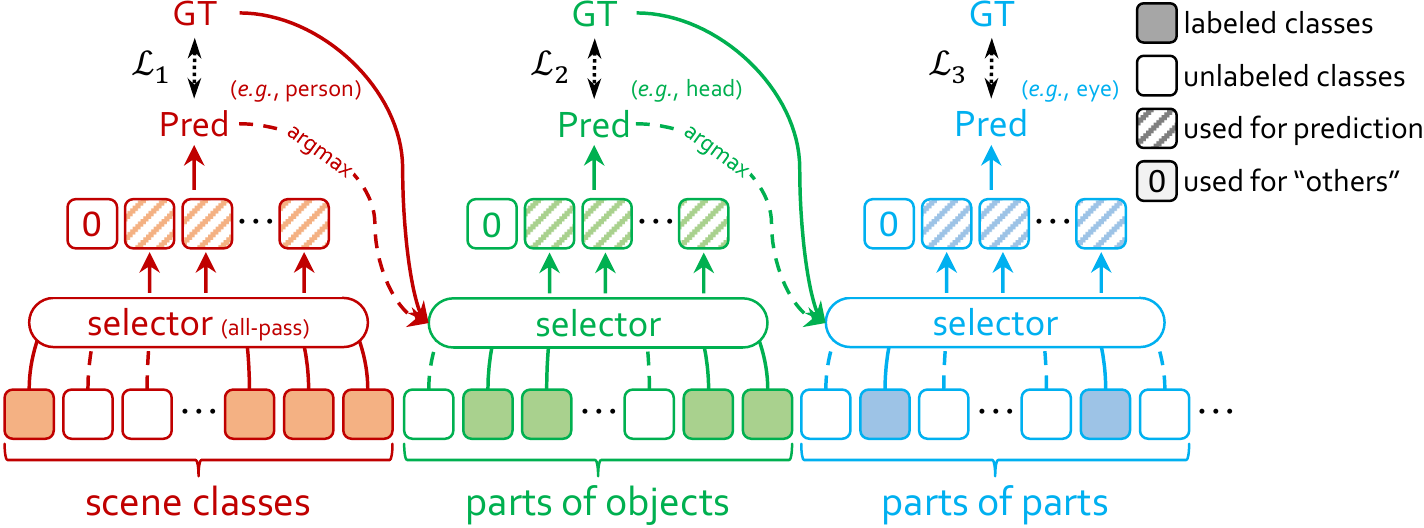}
\vspace{-0.6cm}
\caption{An illustration of how Type-I requests are processed for \textit{one pixel}. Each slot indicates the predicted logit (normalized inner-product) of one class. The \textit{selector} is used to look up the knowledge base for child nodes of current class (decided by ground-truth and upper-level prediction during training and testing, respectively). Only labeled classes are used for computing loss and different losses are added together. In the testing stage, dashed lines are activated and the ground-truth is not accessible.}
\label{fig:group_loss}
\end{figure}

\noindent
\textbf{Implementation and training details.}\quad
We implement the method with two individual models, and use a pre-trained and frozen CLIP~\cite{radford2021learning} model to generate text embeddings. \textbf{(1)} For Type-I requests, we adopt a prevailing Transformer-based model, SegFormer~\cite{xie2021segformer} with its customized backbone MiT-B0/B5, to extract pixel-wise visual features. For efficient training, requests of all object and part classes are processed \textit{in parallel}. For each visual feature vector, we compute the inner-product between it and all text embeddings. Then, the softmax with cross-entropy loss is computed. Note that only the sub-classes that appear in the knowledge base and are also labeled in the current instance are considered -- other classes are ignored because it is unclear if they appear but are not annotated. A pixel may have multiple labels corresponding to different semantic levels, and thus multiple loss terms may be computed, as shown in Figure~\ref{fig:group_loss}. This strategy is crucial for learning from highly incomplete annotations (see Section~\ref{experiments:ade}). \textbf{(2)} For Type-II requests, we adopt CondInst~\cite{tian2020conditional} where all positive feature locations are viewed as probes and trained \textit{in parallel} for each image. Note that in \name, the probes are allowed to lie anywhere on the instance, so we modify CondInst to sample positive training locations from the entire instance mask, instead of the central region~\cite{tian2019fcos} of instance. This training trick, named \textit{mask sampling}, improves instance segmentation accuracy in our setting. For further implementation details, please refer to Appendix~\ref{details}.

\section{Experiments}
\label{experiments}

\subsection{Datasets and Metric}
\label{experiments:evaluation}

We investigate \name on two segmentation datasets with whole-part hierarchies. \textbf{(1)} The Cityscapes-Panoptic-Parts (CPP) dataset~\cite{degeus2021part} inherited the images and $19$ object classes from Cityscapes~\cite{cordts2016cityscapes}, and further annotated two groups of part classes for \textsf{vehicles} and \textsf{persons}. As a result, there are $5$ object classes having parts, with $9$ non-duplicate part classes in total. There is a high ratio of instances annotated with parts, so it is relatively easy for the conventional approaches to make use of this dataset. \textbf{(2)} The ADE20K dataset~\cite{zhou2019semantic} has a much larger number of object and part classes. We follow the convention to use the $150$ most frequent scene classes for top-level semantic segmentation, and $100$ countable objects out of them are used for instance segmentation. Besides, for finer-grained whole-to-part segmentation, we select $82$ part classes (non-duplicate) belonging to $40$ instance classes, and $29$ part-of-parts classes belonging to $17$ upper-level part classes. Please refer to Appendix~\ref{ade20k:preparation} for the details of data preparation. ADE20K has highly sparse and incomplete annotations of parts (and parts of parts), raising difficulties for training and evaluation -- to the best of our knowledge, conventional approaches have not yet quantitatively reported part-based segmentation results on this dataset.

To measure the segmentation quality, we design a metric named hierarchical panoptic quality (HPQ) which can measure the accuracy of a recognition tree of \textit{any depth}. HPQ is a recursive metric. For a leaf node, HPQ equals to mask IoU, otherwise, we first compute the class-wise HPQ:
\begin{equation}
\label{eqn:hpq}
\mathrm{HPQ}_c\!\left(\mathbf{t}_l\right)=\frac{\sum_{\mathbf{t}_{l'}\in\mathcal{TP}_c\cap\mathcal{U}_l}\mathrm{HPQ}\!\left(\mathbf{t}_{l'}\right)}{\left|\mathcal{TP}_c\cap\mathcal{U}_l\right|+\frac{1}{2}\left|\mathcal{FP}_c\cap\mathcal{U}_l\right|+\frac{1}{2}\left|\mathcal{FN}_c\cap\mathcal{U}_l\right|},
\end{equation}
where $\mathcal{TP}_c$, $\mathcal{FP}_c$, and $\mathcal{FN}_c$ denote the sets of true-positives, false-positives, and false-negatives of the $c$-th class, respectively. The true-positive criterion is $\mathrm{HPQ}_c\!\left(\mathbf{t}_{l'}\right)\geqslant0.5$. The HPQ values of all active classes  $\mathrm{HPQ}_c\!\left(\mathbf{t}_0\right)$ are averaged into $\mathrm{HPQ}\!\left(\mathbf{t}_0\right)$ at the root node. HPQ is related to prior metrics, \textit{e.g.}, it degenerates to the original PQ~\cite{kirillov2019panoptic} if there is no object-part hierarchy, and gets similar to PartPQ~\cite{degeus2021part} if the object-part relationship is one-level (\textit{i.e.}, parts cannot have sub-parts) and parts are only semantically labeled (\textit{i.e.}, no part instances) -- this is the case of the CPP dataset (see Appendix~\ref{cpp:metric}). Yet, HPQ is easily generalized to more complex hierarchies.

\begin{table}[!t]
\centering
\caption{Individual segmentation accuracy on CPP. \textbf{Left}: Semantic segmentation (Type-I) on Level-1 (\textit{i.e.}, scene-level, \textit{e.g.}, \textsf{car}) and Level-2 (\textit{i.e.}, part-level, \textit{e.g.}, \textsf{wheel}) classes. \textbf{Right}: Instance segmentation (Type-II) results on all instances under the non-probing protocol. $^\star$ indicates that BPR~\cite{tang2021look} is used.}

\label{tab:cpp-individual}
\resizebox{\linewidth}{!}{
\renewcommand{\arraystretch}{1.2}
\setlength{\tabcolsep}{0.08cm}
\begin{tabular}{l|cccl|c}
\cline{1-3}\cline{5-6}
\textbf{mIoU} (\%) & \textbf{Lv-1} & \textbf{Lv-2} && \textbf{AP (\%)} & \textbf{Inst.} \\
\cline{1-3}\cline{5-6}
SegFormer (B0)~\cite{xie2021segformer} & 76.54 & -- && CondInst~\cite{tian2020conditional} (R50) & 36.6 \\
w/ CLIP & 77.35 & -- && w/ CLIP & 36.8 \\
w/ CLIP \& parts (ours) & \textbf{77.39} & \textbf{75.42} && w/ CLIP \& mask samp. & \textbf{37.8} \\
\cline{1-3}\cline{5-6}
SegFormer (B5)~\cite{xie2021segformer} & 82.25 & -- && CondInst~\cite{tian2020conditional} (R50)$^\star$ & 39.2 \\
w/ CLIP & \textbf{82.40} & -- && w/ CLIP & 39.6 \\
w/ CLIP \& parts (ours) & 82.33 & \textbf{78.48} && w/ CLIP \& mask samp. & \textbf{40.5} \\
\cline{1-3}\cline{5-6}
\end{tabular}}
\end{table}

\subsection{Results on CPP}
\label{experiments:cpp}

\noindent
\textbf{Individual tests.}\quad In CPP, a dataset with relatively complete part annotations, we evaluate Type-I and Type-II requests individually. Results are shown in Table~\ref{tab:cpp-individual}. For \text{Type-I} requests (\textit{i.e.}, semantic segmentation), introducing text embedding as flexible classes improves the accuracy slightly, and training together with part-level classes also produces reasonable results. That said, language-based segmentation models can handle non-part and part classes simultaneously. For \text{Type-II} requests (\textit{i.e.}, instance segmentation), to fairly compete with other methods, we do not use hand-crafted probes in the testing stage. Hence, we densely sample probes on the semantic segmentation regions, each of which generating a candidate instance, and filter the results using non-maximum suppression (NMS) (see Appendix~\ref{details:nonprobing} for details). This is called the non-probing-based setting, which will be used throughout this section, and we also study a more flexible probing-based setting (requires user interaction) in Appendix~\ref{details:probing}. As shown in Table~\ref{tab:cpp-individual}, we achieve higher instance segmentation accuracy. A useful training trick is mask sampling, which makes the model insensitive to the position of probes, as elaborated in Appendix~\ref{details:probing}.


\begin{table}[!t]
\centering
\caption{PartPQ on all, NP (\textit{i.e.}, without parts), and P (\textit{i.e.}, with parts) classes on CPP. $^\star$ indicates that instance segmentation is improved by BPR~\cite{tang2021look}. We also show HPQ values of our method but they cannot be compared against prior methods.}
\label{tab:cpp-comparison}
\resizebox{\linewidth}{!}{
\renewcommand{\arraystretch}{1.2}
\setlength{\tabcolsep}{0.08cm}
\begin{tabular}{l|ccc}
\hline
\textbf{PartPQ (\%)} & \textbf{All} & \textbf{NP} & \textbf{P} \\
\hline
UPSNet + DeepLabv3+~\cite{degeus2021part} & 55.1 & 59.7 & 42.3 \\
HRNet-OCR + PolyTransform + BSANet~\cite{degeus2021part} & 61.4 & 67.0 & 45.8 \\
Panoptic-PartFormer (ResNet50)~\cite{li2022panoptic} & 57.4 & 62.2 & 43.9 \\
Panoptic-PartFormer (Swin-base)~\cite{li2022panoptic} & 61.9 & \textbf{68.0} & 45.6 \\ \hline
\name: SegFormer (B0) + CondInst (R50) & 57.1 & 61.6 & 44.2 \\
\quad \textcolor{gray}{\textit{evaluated using HPQ (\%)}} & \textcolor{gray}{\textit{56.0}} & \textcolor{gray}{\textit{60.5}} & \textcolor{gray}{\textit{43.3}} \\
\name: SegFormer (B5) + CondInst (R50)$^\star$ & \textbf{62.5} & 67.5 & \textbf{48.6} \\
\quad \textcolor{gray}{\textit{evaluated using HPQ (\%)}} & \textcolor{gray}{\textit{61.6}} & \textcolor{gray}{\textit{66.6}} & \textcolor{gray}{\textit{47.4}} \\
\hline
\end{tabular}}
\end{table}

\noindent
\textbf{Comparison to recent approaches.}\quad We combine the best practice into overall segmentation. Results are shown in Table~\ref{tab:cpp-comparison}. We compute both the PartPQ and HPQ metrics and use PartPQ to compare against previous approaches. Note that computing PartPQ is only possible on two-level datasets such as CPP. As shown, our method reports competitive accuracy among recent works~\cite{degeus2021part,li2022panoptic}.
In addition, when we evaluate NP (without defined parts) and P (with parts) classes separately, we find that \name has clear advantages in the latter set, validating the benefit of step-wise visual recognition, \textit{i.e.}, the model can recognizing objects first and then parts. \textit{Note that this work is actually not optimized for higher results on the fully-annotated CPP dataset, the most important advantages are the abilities to learn complex hierarchies from incomplete annotations and adapt to new concepts easily, as investigated below.}

\begin{table}[!t]
\centering
\caption{Segmentation accuracy of using different ratios of annotations. \textbf{Left}: Setting (1) in which all semantic/instance and a portion of part annotations are preserved. \textbf{Right}: Setting (2) in which both semantic and part annotations are preserved to the corresponding ratio. See the main texts for details.}
\vspace{-0.2cm}
\label{tab:cpp-incomplete}
\resizebox{\linewidth}{!}{
\renewcommand{\arraystretch}{1.2}
\setlength{\tabcolsep}{0.12cm}
\begin{tabular}{l|cc|ccl|cc|c}
\cline{1-4}\cline{6-9}
\multirow{2}{*}{\textbf{Ratio}} & \multicolumn{2}{c|}{\textbf{mIoU (\%)}} & \textbf{HPQ} && \multirow{2}{*}{\textbf{Ratio}} & \multicolumn{2}{c|}{\textbf{mIoU (\%)}} & \textbf{HPQ} \\
& \textbf{Lv-1} & \textbf{Lv-2} & \textbf{(\%)} && & \textbf{Lv-1} & \textbf{Lv-2} & \textbf{(\%)} \\
\cline{1-4}\cline{6-9}
100\% & 82.33 & 78.48 & 60.2 && 100\% & 82.33 & 78.48 & 60.2 \\
50\% & 81.82 & 77.59 & 59.8 && 75\% & 81.49 & 77.81 & 59.3 \\
30\% & 81.99 & 76.52 & 59.4 && 50\% & 79.01 & 76.23 & 56.9 \\
15\% & 81.72 & 65.51 & 57.3 && 30\% & 73.53 & 69.72 & 51.7 \\ 
\cline{1-4}\cline{6-9}
\end{tabular}}
\end{table}

\noindent
\textbf{Learning from incomplete annotations.}\quad
We have discussed how \name deals with incomplete annotations in Section~\ref{problem:advantages} and Section~\ref{approach}. Here, we study this mechanism by performing two experiments on CPP. \textbf{(1)} We preserve all semantic and instance annotations but randomly choose a certain portion of part annotations. \textbf{(2)} Beyond (1), we further remove part of semantic annotations -- note that this requires associating each image with an individual knowledge base. Please refer to Appendix~\ref{cpp:partsampling} for details. Results are shown in Table~\ref{tab:cpp-incomplete}. We find that \name, without any modification, adapts to both scenarios easily. This property largely benefits our investigation on the ADE20K dataset, where a considerable portion of part annotations is missing and there is an eager call for few-shot incremental learning.

\begin{figure*}
\vspace{-0.6cm}
\centering
\includegraphics[width=\linewidth]{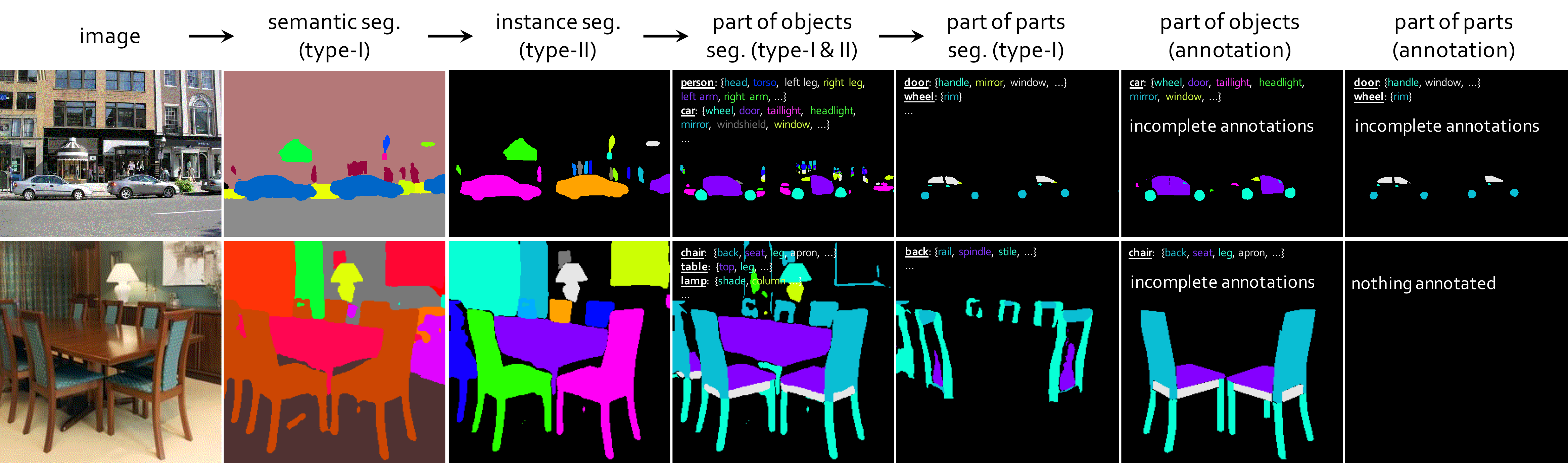}
\vspace{-0.6cm}
\caption{Visualization results of \name on ADE20K. Black areas in prediction indicate the \textsf{others} (\textit{i.e.}, unknown) class. The corresponding sub-knowledge is listed in the blank area for reference. \textit{Best viewed in color and by zooming in for details.}}
\label{fig:visualization}
\end{figure*}

\subsection{Results on ADE20K}
\label{experiments:ade}

\noindent
\textbf{Adapting conventional methods to ADE20K.}\quad
We first note that the annotations on ADE20K are highly incomplete, making it difficult for the conventional segmentation methods to learn the complex, hierarchical concepts. We conjecture it to be the main reason that no prior works have ever reported quantitative results for part-aware segmentation on ADE20K. Conventional solutions usually adopt two ways to handle unlabeled pixels, either ignoring all of them or assigning them to an extra background class. We argue that the latter solution is prone to introducing erroneous information because it cannot distinguish undefined parts from unlabeled parts. For comparison, we establish a competitor method (which we refer to as \textit{convention}) in which all three levels (\textit{i.e.}, objects, object parts, parts of parts) are jointly trained in a multi-task manner, following the former conventional solution to ignore all unlabeled pixels. Meanwhile, \name is easily adapted to this scenario. We simply inherit the best practice from the CPP experiments. The segmentation models (SegFormer and CondInst) and training epochs remain the same for fair comparison. That said, the main difference between \name and the competitor lies in the way of organizing training data to ease learning multi-level hierarchies from incomplete annotations.

\begin{table}[!t]
\centering
\caption{Comparison of HPQ on all, NP (\textit{i.e.}, without parts), and P (\textit{i.e.}, with parts defined) classes on ADE20K. Both methods use CondInst (R50) for instance segmentation. P$^\dagger$ indicates that instances without part annotations are ignored, so that the metric focuses more on the quality of part-level segmentation.}
\vspace{-0.2cm}
\label{tab:ade-comparison}
\resizebox{\linewidth}{!}{
\renewcommand{\arraystretch}{1.2}
\setlength{\tabcolsep}{0.2cm}
\begin{tabular}{ll|cccc}
\hline
\multicolumn{2}{c|}{\textbf{HPQ (\%)}} & \textbf{All} & \textbf{NP} & \textbf{P} & \textbf{P$^\dagger$} \\
\hline
\multicolumn{1}{c|}{Convention} & \multirow{2}{*}{SegFormer-B0} & 25.6 & 24.9 & 27.7 & 16.8 \\
\multicolumn{1}{c|}{\text{\name}} & & \textbf{27.2} & \textbf{25.7} & \textbf{31.1} & \textbf{21.9} \\
\hline
\multicolumn{1}{c|}{Convention} & \multirow{2}{*}{SegFormer-B5} & 31.8 & 31.8 & 32.0 & 18.9 \\
\multicolumn{1}{c|}{\text{\name}} & & \textbf{33.9} & \textbf{32.8} & \textbf{36.9} & \textbf{27.1} \\
\hline
\end{tabular}}
\end{table}

\begin{figure}[!t]
\vspace{-0.6cm}
\centering
\includegraphics[width=\linewidth]{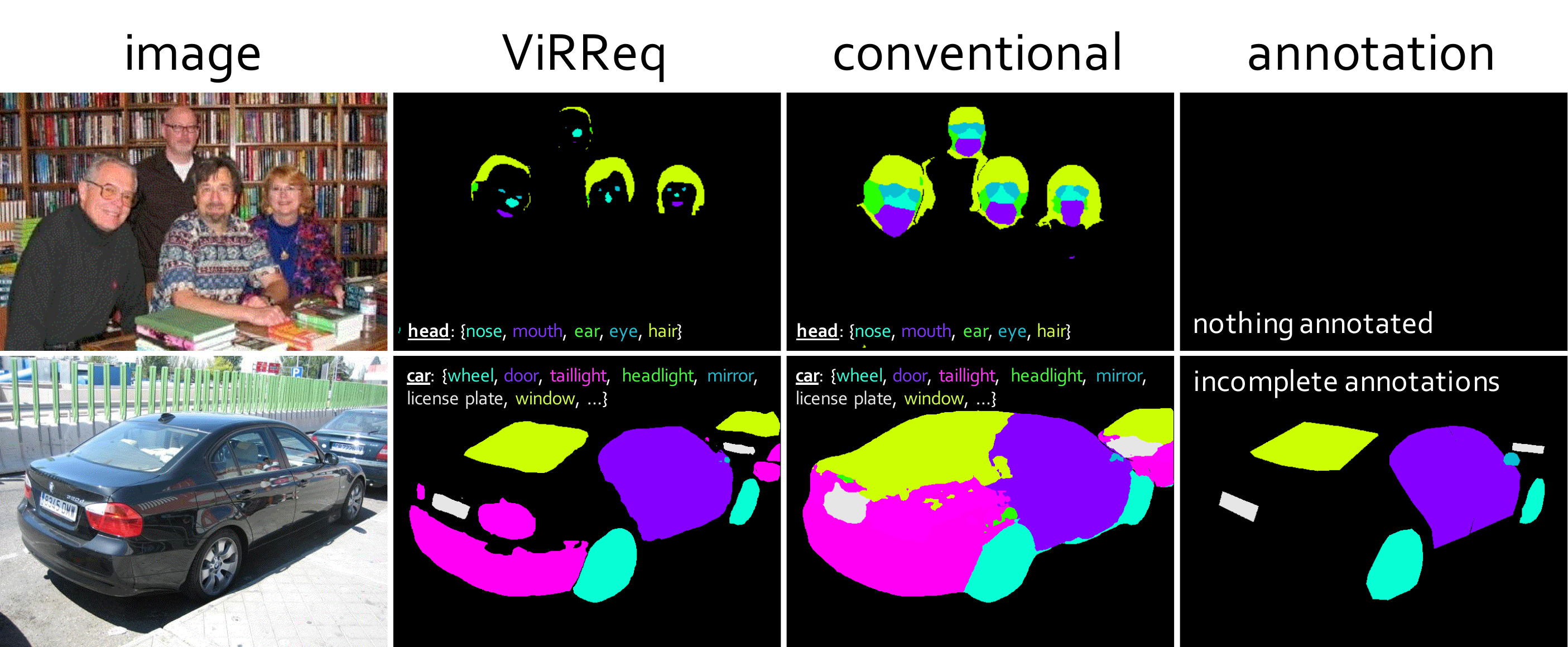}
\vspace{-0.6cm}
\caption{Qualitative comparison between \name and convention. Note how convention over-estimate the boundary of parts.}
\label{fig:compare_to_baseline}
\end{figure}

\noindent
\textbf{Quantitative comparison.}\quad
Results are shown in Table~\ref{tab:ade-comparison}. We report the HPQ values for different subsets of concepts. Without complex adaptation, \name surpasses the convention, showing the ability of \name in learning from highly incomplete annotations. We expect the improvement to be larger when more and more parts or parts of parts are introduced -- in such scenarios, \name enjoys the ability of decomposing recognition into requests which gets rid of the burden of distinguishing from a large number of part classes. \name offers a benchmark in this direction and offers room of improvement for future research.

\noindent
\textbf{Qualitative comparison.}\quad
We visualize some segmentation results of \name in Figure~\ref{fig:visualization} to show its ability of recognizing complex whole-part hierarchies. Moreover, we compare \name to the competitor in learning from incomplete part annotations. As shown in Figure~\ref{fig:compare_to_baseline}, \name produces much cleaner segmentation results in which most undefined pixels are correctly recognized as \textsf{others}.

\begin{figure}[!t]
\vspace{-0.6cm}
\centering
\includegraphics[width=\linewidth]{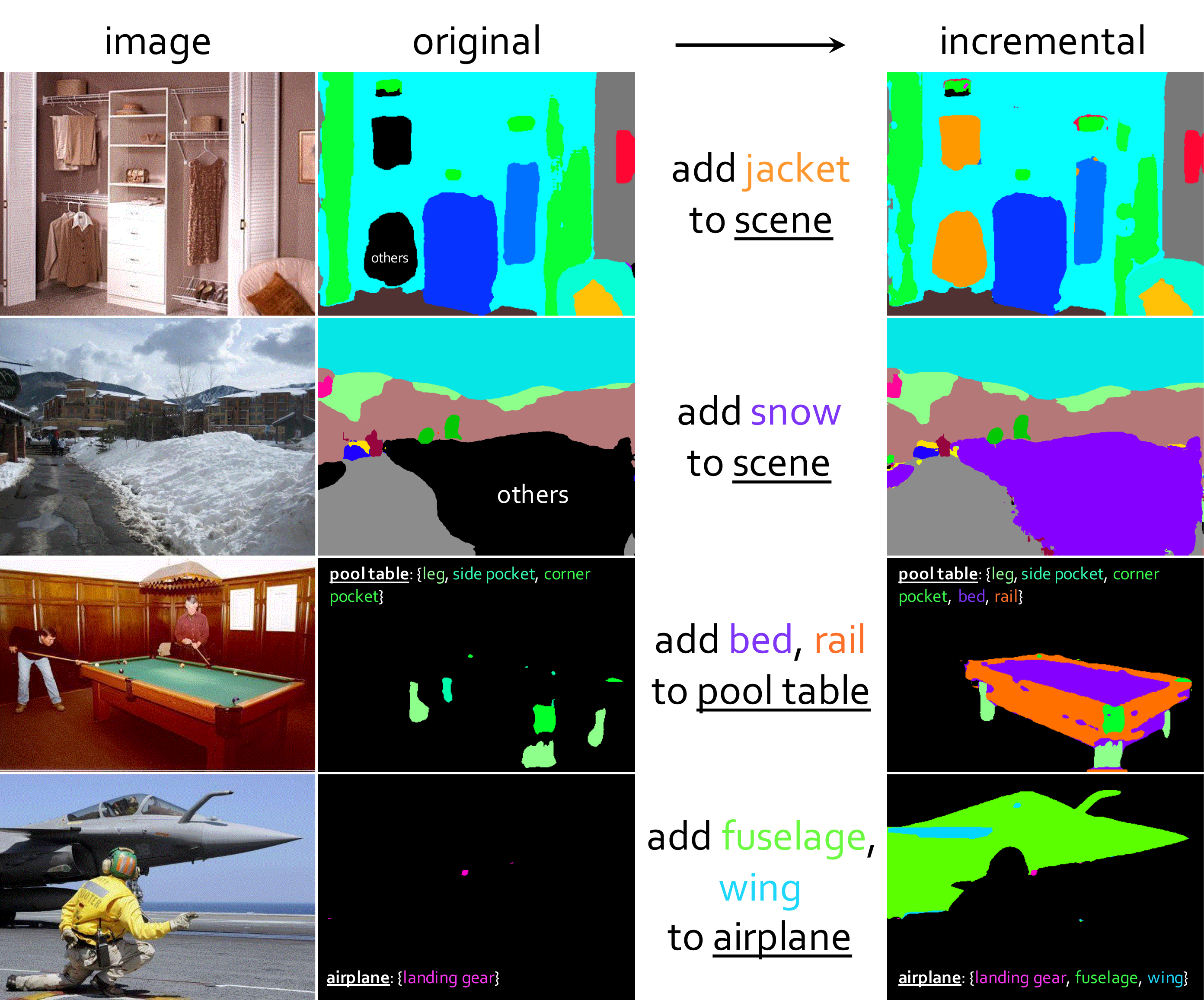}
\vspace{-0.6cm}
\caption{Qualitative results of few-shot incremental learning on object parts (top two rows), and parts of parts (bottom two rows). Prior to incremental learning, these new concepts are recognized as \textsf{others} (the black areas) which is as expected.}
\label{fig:visualization_incremental}
\end{figure}

\noindent
\textbf{Few-shot incremental learning.}\quad
We add $50$ top-level objects and $19$ parts to the original knowledge base built in ADE20K. For each new concept, $20$ instances are labeled, yet the existing data may contain unlabeled instances of these classes. We mix old and new data and fine-tune the trained SegFormer-B5 for $1/4$ of base training iterations. We follow CLIP-Adapter~\cite{gao2021clip} to integrate an additional bottleneck layer for feature alignment, which is the only layer that gets updated during the fine-tuning. More details on data preparation and implementation are provided in Appendix~\ref{more_on_incremental}. We show qualitative results in Figure~\ref{fig:visualization_incremental}, where \name easily absorbs new classes into the knowledge base meanwhile the base classes remain mostly unaffected. We emphasize that the incremental learning ability comes from (i) the query-based, language-driven recognition framework and (ii) data versioning (see Section~\ref{problem:advantages}).

\begin{figure}[!t]
\vspace{-0.2cm}
\centering
\includegraphics[width=\linewidth]{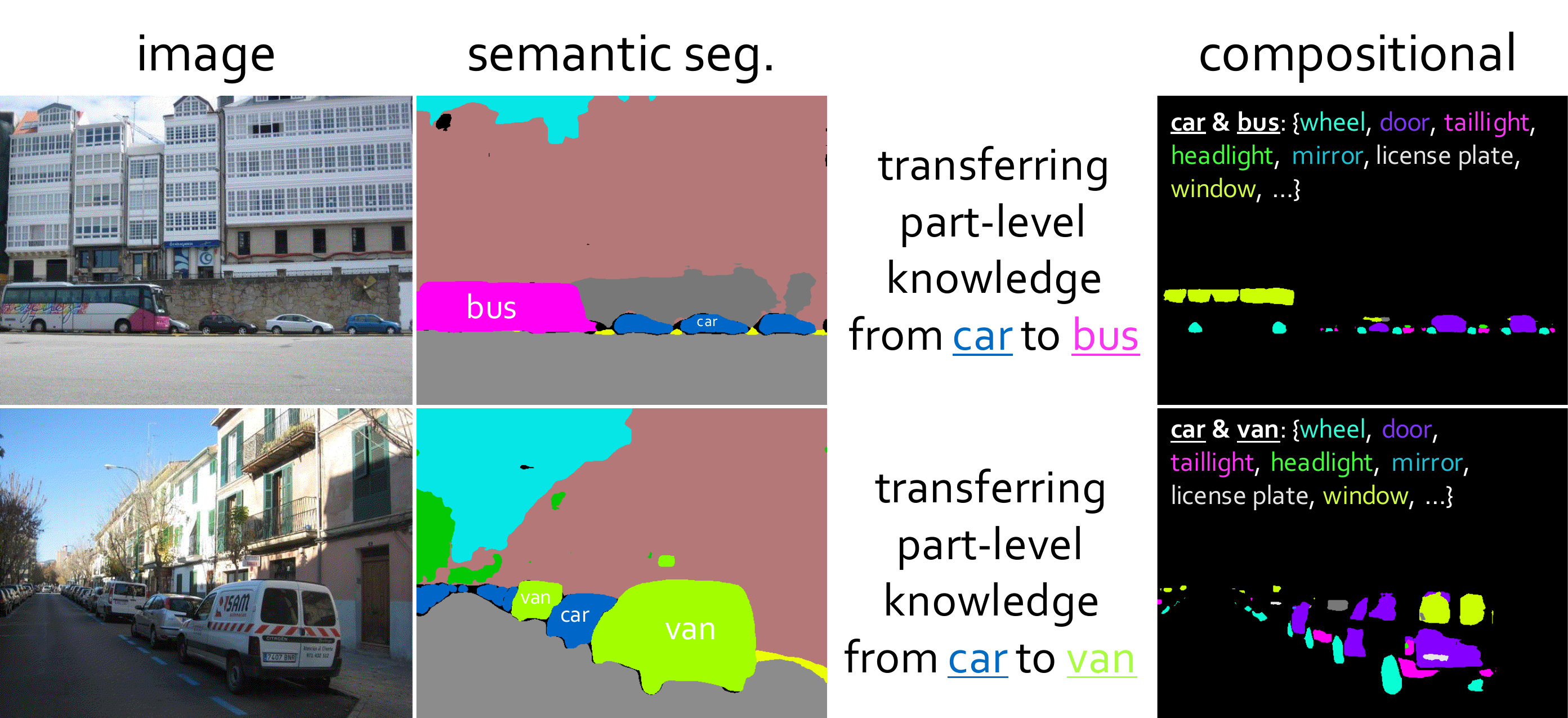}
\caption{Visualization of compositional segmentation. Note that the base class (\textsf{car}) and its parts are unaffected.}
\vspace{-0.6cm}
\label{fig:visualization_compositional}
\end{figure}

\noindent
\textbf{Open-domain recognition.}\quad
We investigate two interesting scenarios of open-domain recognition. The first one is anomaly segmentation, \textit{i.e.}, finding unknown concepts in an image, which is naturally supported in \name due to the existence of the \textsf{others} class in each sub-task. Object-level and part-level examples are shown in Figures~\ref{fig:compare_to_baseline} and~\ref{fig:visualization_incremental}, respectively. Note that \name can easily convert unknown classes to known classes via incremental learning. The second one involves compositional segmentation, \textit{i.e.}, transferring part-level knowledge from one class (\textit{e.g.}, \textsf{car}) to others (\textit{e.g.}, \textsf{bus} or \textsf{van}) without annotating new data but directly copying the sub-knowledge from the old class to new classes. As shown in Figure~\ref{fig:visualization_compositional}, \name learns these parts on new classes without harming the old class or its parts.

\section{Conclusions and Future Directions}
\label{conclusions}

In this paper, we present \text{\name}, a novel setting that pushes visual recognition towards unlimited granularity. The core idea is to decompose the end-to-end recognition task into requests, each of which performs a single step of recognition, hence alleviating the conflict between granularity and certainty. We offer a simple baseline that uses language as class index, and achieves satisfying segmentation accuracy on two segmentation datasets with multiple whole-part hierarchies. In particular, thanks to the ability of learning from incomplete annotations, we report part-aware segmentation accuracy (using HPQ, a newly defined metric for \name) on ADE20K for the first time.

From this preliminary work, we learn the lesson that visual recognition has some important yet unsolved issues (\textit{e.g.}, unlimited granularity). Language can assist the definition (\textit{e.g.}, using text-based class indices for open-domain recognition), but solving the essential difficulty requires insights from vision itself. From another perspective, the baseline can be seen as pre-training a language-assisted vision backbone and querying it using vision-friendly prompts (\textit{i.e.}, Type-I/II requests). In the future, we will continue with this paradigm to explore the possibility of unifying various visual recognition tasks, for which two promising directions emerge: (i) designing an automatic method for learning and updating the knowledge base from training data, and (ii) closing the gap between upstream pre-training and downstream fine-tuning with better prompts.

{\small
\bibliographystyle{ieee_fullname}
\bibliography{egbib}
}

\clearpage
\appendix

\section{Implementation Details}
\label{details}

\subsection{Model Implementation}
\label{details:architecture}

\noindent
\textbf{Text encoder.}\quad We adopt the Contrastive Language–Image Pre-training (CLIP)~\cite{radford2021learning} model as the text encoder throughout, which supports variable-length text labels by design. The pre-trained and frozen CLIP-ViT-B/32 model was used, which produces a $512$-dimensional embedding vector for each text label. In this paper, we simply use the class names (mostly containing 1 or 2 words, \textit{e.g.}, \textsf{person}, \textsf{traffic sign}) defined in the datasets.

\begin{figure}[!t]
\centering
\includegraphics[width=\linewidth]{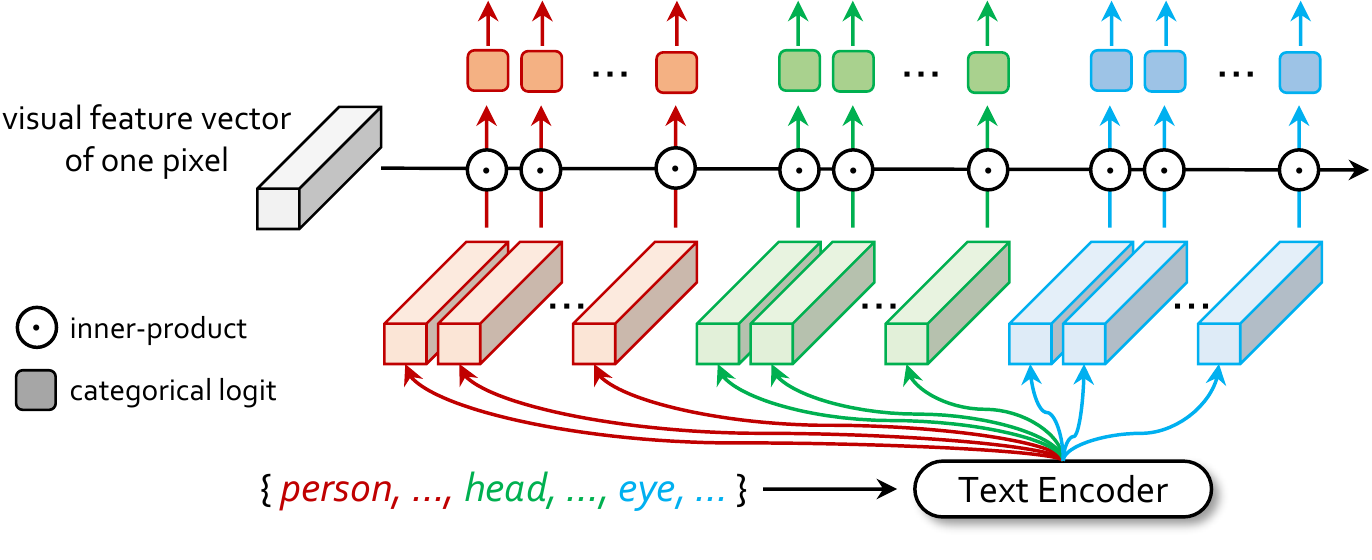}
\vspace{-0.8cm}
\caption{An illustration of generating language-based queries and performing vision-language interaction for Type-I requests (the case of \textit{one pixel} is plotted for simplicity). The visual feature vector and the text embedding have the same dimension (\textit{e.g.}, $512$). The input text labels can be freely replaced by any texts of arbitrary length. See the main texts for more details.}
\label{fig:appendix-inner-product}
\end{figure}

\noindent
\textbf{Type-I (SegFormer).}\quad We adopt the SegFormer~\cite{xie2021segformer} implementation in the \texttt{MMSegmentation}~\cite{mmseg2020} codebase to perform whole-to-part segmentation for Type-I requests. Nota that SegFormer can be freely replaced by any other visual feature extractors.  We additionally provide an illustration of generating language-based queries and performing vision-language interaction for Type-I requests, as shown in Figure~\ref{fig:appendix-inner-product}. SegFormer uses a series of Mix Transformer encoders (MiT) with different sizes as the visual backbones. In this paper, we mainly adopt the lightweight model (MiT-B0) for fast development and the largest model (MiT-B5) towards better performance. The output feature maps have $512$ channels, which can be directly interacted with the text embedding vectors. We additionally apply a projection module (four fully-connected layers) on the text features for better feature alignment. Note that the categorical logits after vision-language feature interaction is usually divided by a temperature parameter $\tau$, $\mathbf{u}_{w,h}=(\mathbf{E}^\top\cdot\mathbf{f}_{w,h})/\tau$, where $\tau$ controls the concentration level of the distribution~\cite{radford2021learning,li2022language}. We follow CLIP to set $\tau$ as a learnable parameter, which is initialize to be $0.07$ in the start of the training stage.

\begin{figure}[!t]
\centering
\includegraphics[width=\linewidth]{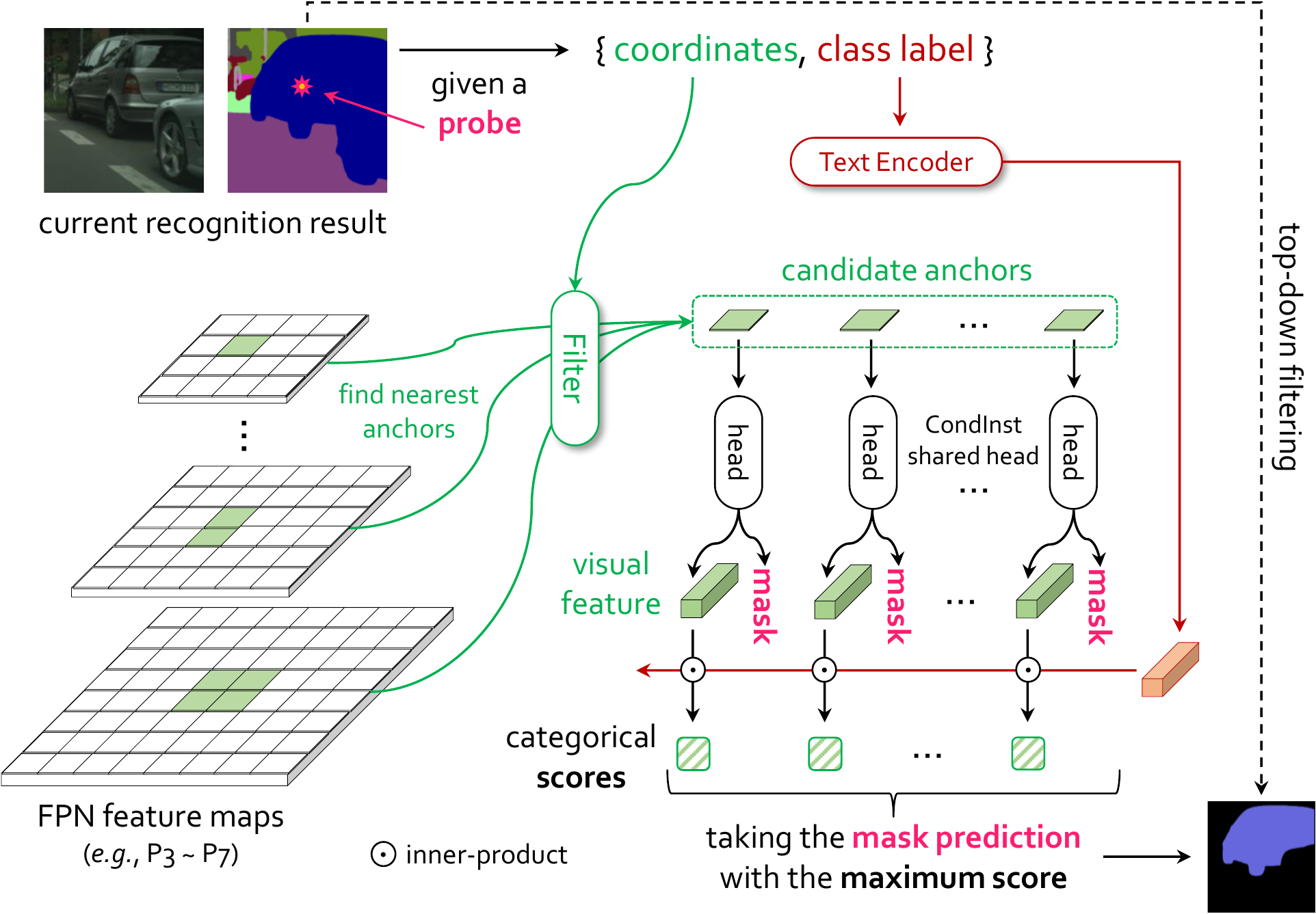}
\vspace{-0.6cm}
\caption{An illustration of how Type-II requests are processed with CondInst. The probe is sampled (or clicked by user) from the semantic region of current recognition result, aiming at segmenting the instance that occupies this probe. The \textit{filter} is used to select the spatially related anchors (or named proposals) by the positional embedding $\mathbf{p}_k$ (equivalents to probe coordinates in our implementation), and the  preserved anchors are used for subsequent prediction. Prediction with the highest categorical score (obtained by inner-product with text embedding) is chosen as the final result. See the main texts for more details.}
\label{fig:appendix-condinst}
\end{figure}

\noindent
\textbf{Type-II (CondInst).}\quad We adopt the CondInst~\cite{tian2020conditional} implementation in the \texttt{AdelaiDet}~\cite{tian2019adelaidet} codebase to perform instance segmentation. Nota that CondInst can also be replaced by other instance segmentation models (\textit{e.g.}, Mask R-CNN~\cite{he2017mask} or SOLOv2~\cite{wang2020solov2}) with marginal modification. CondInst treats all feature locations on multiple feature pyramids as instance proposals, which is naturally compatible with the design of the probing-based inference. We additionally provide an illustration of how Type-II requests are processed with CondInst, as shown in Figure~\ref{fig:appendix-condinst}. Since the output feature maps (from the last layer of classification branch) have $256$ channels, we apply a linear projection to transform the dimension of text embedding from $512$ to $256$ to perform feature interaction. We mainly adopt ResNet-50 as the visual backbone of CondInst. We empirically observed that the results were improved slightly ($\sim0.3\%$ mAP) with a larger backbone, ResNet-101 -- we conjecture that the limited improvement lies in the small dataset size of CPP and ADE20K. In addition, we observed that the instance segmentation model is more sensitive to the choice of $\tau$. For example, using $\tau=0.07$ leads to sub-optimal results. We empirically found that setting $\tau=1.0$ and adding a learnable bias term (partially following GLIP~\cite{li2022grounded}) works consistently better.

\subsection{Model Optimization}
\label{details:optimization}

For Type-I requests, we almost follow the same training protocol as SegFormer, except that classes of different semantic levels (\textit{e.g.}, objects, object parts, parts of parts) are jointly trained in one model (see Figure~\ref{fig:group_loss}). The MiT encoder was pre-trained on Imagenet-1K and the decoder was randomly initialized. For data augmentation, random resizing with ratio of $0.5\sim2.0$, random horizontal flipping, and random cropping ($1024\times1024$ for CPP and $512\times512$ for ADE20K) were applied. The model was trained with AdamW optimizer for $160K$ iterations on $8$ NVIDIA Tesla V100 GPUs. The initial learning rate is $0.00006$ and decayed using the poly learning rate policy with the power of $1.0$. The batch size is $8$ for CPP and $16$ for ADE20K.

For Type-II requests, since CondInst~\cite{tian2020conditional} have not reported results on Cityscapes and ADE20K, we implemented by ourselves. Due to the small dataset size, we initialized the model with the MS-COCO~\cite{lin2014microsoft} pre-trained CondInst, which increases the results by about $1\%$ AP. The model was trained with SGD optimizer on $8$ NVIDIA Tesla V100 GPUs. For CPP, we used the same training configurations as Mask R-CNN on Cityscapes~\cite{wu2019detectron2} and the model was trained for $24K$ iterations with the batch size of $8$. For ADE20K, the maximum number of proposals sampled during training was increased from $500$ to $1\rm{,}000$ since more instances appeared in an image than MS-COCO. The model was trained for $40K$ iterations with the initial learning rate being $0.01$ and decayed at $30K$ iterations. During training, random resizing with ratio of $0.5\sim2.0$ and random cropping ($640\times640$) were applied. Other configurations are the same as the original CondInst.

\subsection{Non-Probing Segmentation}
\label{details:nonprobing}

\begin{figure}[!t]
\centering
\includegraphics[width=\linewidth]{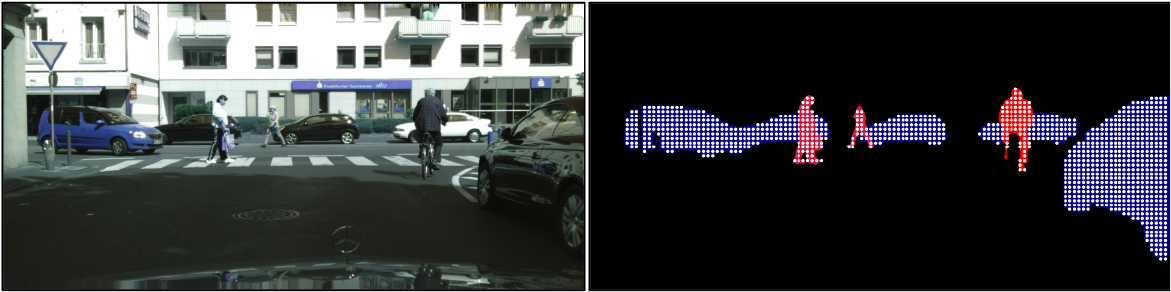}
\vspace{-0.6cm}
\caption{Illustration of sampling probes with \textit{stride} (\textit{e.g.}, $16$).}
\label{fig:appendix-stride}
\end{figure}

\begin{table}[!t]
\caption{Non-probing-based instance segmentation results w.r.t. different \textit{strides}.}
\vspace{-0.4cm}
\label{tab:cpp-instance-stride}
\vspace{0.2cm}
\centering
\setlength{\tabcolsep}{0.18cm}
\renewcommand{\arraystretch}{1.2}
\begin{tabular}{l|cccc}
\hline 
\multirow{2}{*}{\textbf{AP} (\%)} & \multicolumn{4}{c}{\textbf{Non-Probing (w.r.t. \textit{stride})}} \\
& $1$ & $8$ & $16$ & $32$ \\
\hline 
w/o & 37.8 & 37.8 & 36.8 & 33.6  \\
w/ mask samp. & 38.5  & 38.5 & 37.8 & 35.1 \\
\hline 
\end{tabular}
\end{table}

The non-probing-based inference is used to fairly compare against the conventional instance segmentation approaches, \textit{i.e.}, finding all instances \textit{at once}. For this purpose, we densely sample a set of probing pixels based on the semantic segmentation results. Specifically, we regularly sample points with a fixed \textit{stride} in the whole image, and keep the points inside the corresponding semantic regions, as illustrated in Figure~\ref{fig:appendix-stride} (white dots indicate the sampled probes). Each sampled point is viewed as a probing pixel to produce a candidate instance prediction (see Appendix~\ref{details:probing} and Figure~\ref{fig:appendix-condinst} for details). Finally, the results are filtered with NMS (using a threshold of $0.6$ as the same as CondInst). Intuitively, the above procedure can be viewed as an improved CondInst by replacing the builtin classification branch as a standalone pixel-wise classification model. We present the results with respect to different \textit{strides} in Table~\ref{tab:cpp-instance-stride}. As shown, the denser the sampled probes, the higher the mask AP. In addition, enabling mask sampling during training further improves the results (see Appendix~\ref{details:probing} for details of mask sampling). We mainly use the stride of $16$ in experiments.

\subsection{Probing Segmentation}
\label{details:probing}

The probing-based inference, a more flexible setting, is used to simulate the user click to place a probing pixel that lies within the intersection of the predicted semantic region and the ground-truth instance region -- if the intersection is empty, the instance is lost (\textit{i.e.}, IoU is $0\%$). Specifically, we first compute the mass center $(a_{mass}, b_{mass})$ and the bounding box $\mathcal{B}=(a_0,b_0,a_1,b_1)$ of the intersection region. The actual sampling bounding box $\hat{\mathcal{B}}=(\hat{a}_0,\hat{b}_0,\hat{a}_1,\hat{b}_1)$ is determined by a hyper-parameter $\gamma\in\left[0,1\right]$, which controls the centerness of the probing pixels:

\begin{equation}
\begin{aligned}
\hat{a}_0 &= a_{mass} - \gamma(a_{mass} - a_0), \\
\hat{a}_1 &= a_{mass} + \gamma(a_1 - a_{mass}), \\
\hat{b}_0 &= b_{mass} - \gamma(b_{mass} - b_0), \\
\hat{b}_1 &= b_{mass} + \gamma(b_1 - b_{mass}).
\end{aligned}
\end{equation}
The probe is randomly sampled from the intersection of $\hat{\mathcal{B}}$ and the original sampling region. If no candidate pixel lies in that region (since the mass center may outside the instance), the probe is instead sampled from $\hat{\mathcal{B}}$ only. The probe lies exactly on the mass center if $\gamma=0$, and the probe is uniformly distributed on the instance if $\gamma=1$, otherwise, the sampling strategy lies between two extreme situations.

During inference, for each Type-II request we have a triplet $\{a,b,c\}$ (see Section~\ref{approach} for details, the subscript $k$ is omitted here for simplicity). To perform instance segmentation \textit{by request}, we first compute the mapped feature locations $(a_f, b_f) = (\lfloor \frac{a}{s_f} \rfloor, \lfloor \frac{b}{s_f} \rfloor)$ for each FPN level, where $s_f$ is the stride of the $f$-th FPN level, and $s_f \in \{8,16,32,64,128\}$ for CondInst. The five mapped feature locations are viewed as candidates for producing subsequent predictions. Finally, we only choose one prediction with the highest categorical score, where the scores are computed by the inner-product of the pixel-wise feature vectors (extracted from the FPN feature maps) and the text embedding vector of the target class $c$ (generated by the text encoder). This process is illustrated in Figure~\ref{fig:appendix-condinst}. For probing-based inference, there is at most one prediction for an instance, which naturally eliminates the need of non-maximum suppression (NMS).

\begin{figure*}
\centering
\includegraphics[width=16cm]{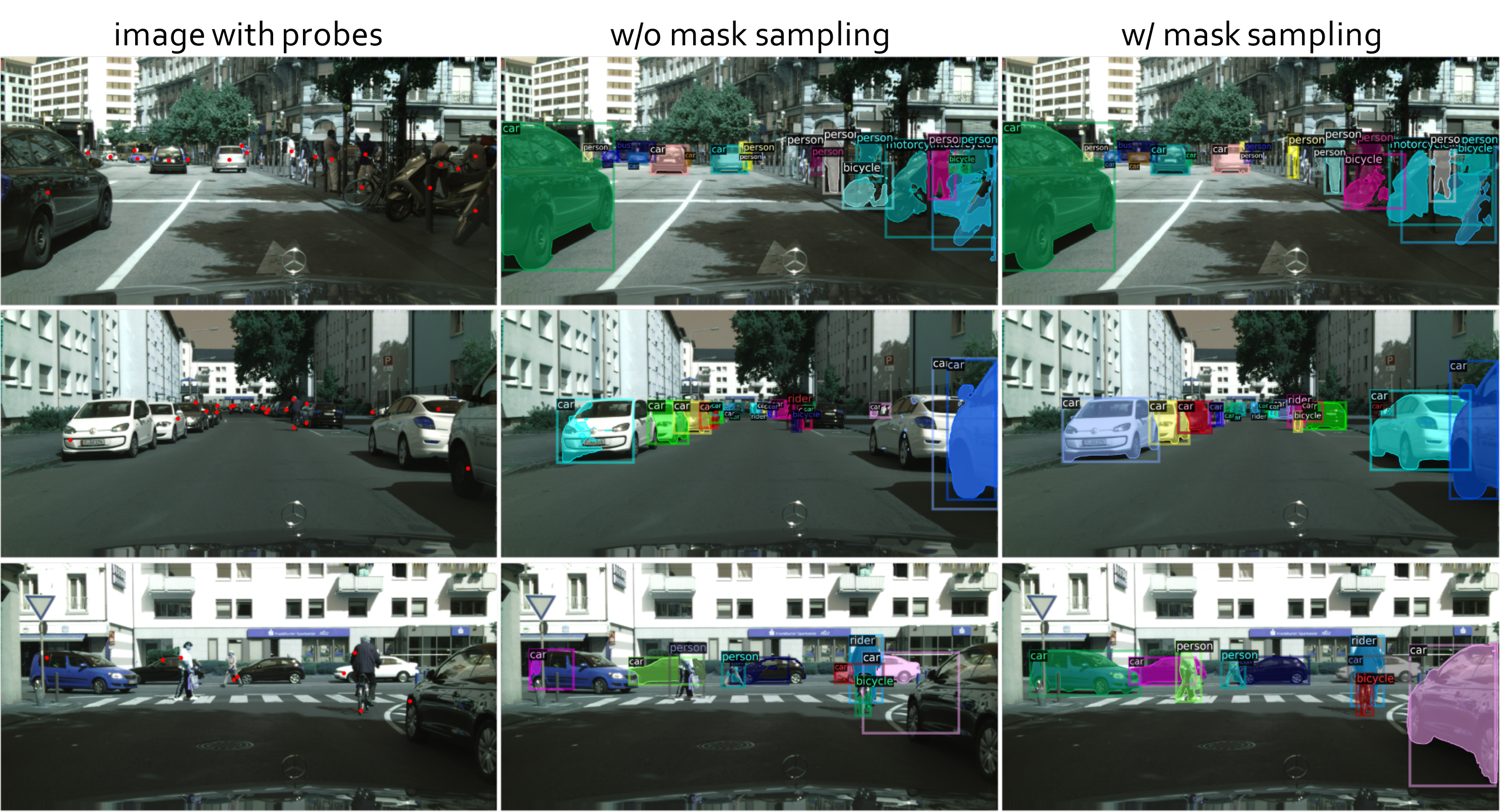}
\caption{Results of using different types of \textcolor{red}{\textit{probes}} (from top to bottom): mass center of instance ($\gamma=0$), random points on the instance ($\gamma=1$), handcrafted low-quality probes that closing to the instance boundaries. \textit{Best viewed digitally with color.}}
\label{fig:appendix-probe-effect}
\end{figure*}

In Table~\ref{tab:appendix-mask-sampling}, we present the mask AP with respect to to different $\gamma$. As shown, AP increases consistently with lower $\gamma$ (\textit{i.e.}, closer to the mass center). We observed that AP decreases dramatically with lager $\gamma$ if the mask sampling strategy was not used during training (the second row). We diagnosed the issue and found that some probes away from the instance center produced unsatisfying results, as shown in Figure~\ref{fig:appendix-probe-effect}. The reason lies in that CondInst only samples positive positions from a small central region of instance by design~\cite{tian2019fcos,tian2020conditional}, thus not all possible probes are properly trained. To address this issue, we instead sample positive positions from the entire ground-truth instance masks. The results are presented in Table~\ref{tab:appendix-mask-sampling} (the last row) and Figure~\ref{fig:appendix-probe-effect} (the last column). As shown, by enabling mask sampling during training, the results are much more robust to the quality of probes. In addition to probing-based inference, we observed that mask sampling is also helpful for non-probing-based inference, as shown in Table~\ref{tab:cpp-individual}.

\begin{table}[!t]
\centering
\caption{Instance segmentation results with non-probing-based and probing-based inference on the CPP dataset. Results are produced using the semantic mask prediction of SegFormer-B5. The hyper-parameter $\gamma$ controls the centerness of the probes.}
\vspace{-0.2cm}
\label{tab:appendix-mask-sampling}
\setlength{\tabcolsep}{0.12cm}
\renewcommand{\arraystretch}{1.2}
\resizebox{\linewidth}{!}{
\begin{tabular}{l|ccccc}
\hline 
\multirow{2}{*}{\textbf{AP} (\%)} & \multirow{2}{*}{\textbf{Non-Probing}} & \multicolumn{4}{c}{\textbf{Probing (w.r.t. $\gamma$)}} \\
& & $0.0$ & $0.2$ & $0.5$ & $1.0$ \\
\hline 
CondInst (R50)~\cite{tian2020conditional} &36.6 & -- & -- & -- & -- \\
w/ CLIP & 36.8 & 39.3 & 39.0 & 34.6 & 24.5 \\
w/ CLIP \& mask samp. & \textbf{37.8} & \textbf{39.4} & \textbf{39.1} & \textbf{37.4} & \textbf{33.5} \\
\hline 
\end{tabular}}
\end{table}

\section{Details of the CPP Experiments}
\label{cpp}

\subsection{Data Statistics}
\label{cpp:statistics}

The Cityscapes Panoptic Parts (CPP) dataset~\cite{degeus2021part} extends the popular Cityscapes~\cite{cordts2016cityscapes} dataset with part-level semantic annotations. There are $9$ part classes belonging to $5$ scene-level classes are annotated for $2\rm{,}975$ training and $500$ validation images in Cityscapes. Specifically, two \textsf{human} classes (person, rider) are labeled with 4 parts (\textit{torso}, \textit{head}, \textit{arm}, \textit{leg}), and three \textsf{vehicle} classes (car, truck, bus) are labeled with 5 parts (\textit{chassis}, \textit{window}, \textit{wheel}, \textit{light}, \textit{license plate}). The CPP dataset provides exhaustive part annotations that each instance mask (belonging to the chosen $5$ classes) is completely partitioned into the corresponding part masks. In our experiments, we use $19$ semantic classes ($8$ out of $19$ are thing classes for instance segmentation) and $9$ non-duplicate part classes in total. Results on the CPP validation set are reported.

\subsection{HPQ vs. PartPQ in CPP}
\label{cpp:metric}

In the scenario of CPP (only two hierarchies), the only difference between PartPQ and HPQ lies in computing the accuracy of the objects that have parts. PartPQ directly averages the mask IoU values of all parts, while HPQ calls for a recursive mechanism. CPP is a two-level dataset (\textit{i.e.}, parts cannot have parts), and all parts are semantically labeled (\textit{i.e.}, no instances are labeled on parts although some of them, \textit{e.g.}, \textsf{wheel} of \textsf{car}, can be labeled at the instance level). In this scenario, (1) the HPQ of a part (as a leaf node) is directly defined as its mask IoU if the corresponding prediction is a true positive (IoU is no smaller than $0.5$) and HPQ equals zero otherwise, and (2) since each part has only one unit, the recognition of each part has either a true positive (IoU is no smaller than $0.5$) or a false positive plus a false negative (IoU is smaller than $0.5$) -- that said, the denominator of HPQ is a constant, equaling to the number of parts. As a result, the values of HPQ are usually lower than PartPQ (see Table~\ref{tab:cpp-comparison}).

\subsection{Sampling incomplete annotations in CPP}
\label{cpp:partsampling}

In Table~\ref{tab:cpp-incomplete}, we report the results of learning from incomplete annotations, where only a subset of annotations are preserved for training. For the setting (1), we preserve all semantic and instance annotations but randomly choose a certain portion of part annotations. Specifically, there are in total $161\rm{,}182$ annotated part-level masks in the CPP training set. We randomly sample a certain ratio (\textit{e.g.}, $15\%\sim50\%$) of part-level masks for training. For the setting (2), we first randomly sample a certain ratio (\textit{e.g.}, $30\%\sim75\%$) of scene-level masks ($34\rm{,}723$ in total), and further sample the same ratio of part-level masks within the preserved scene-level masks, which is a more incomplete scenario. For both settings, evaluation was conducted on the validation set with complete part annotations. We find that ViRReq, without any modification, adapts to both settings easily.

\subsection{More Quantitative Results}
\label{cpp:more_quantitative}

\begin{table}[t]
\caption{Overall segmentation results on CPP, using non-probing-based inference and probing-based inference, respectively. $^\star$ indicates that BPR~\cite{tang2021look} is used.}
\vspace{-0.2cm}
\label{tab:cpp-overall}
\centering
\setlength{\tabcolsep}{0.12cm}
\renewcommand{\arraystretch}{1.2}
\resizebox{0.9\linewidth}{!}{
\begin{tabular}{l|c|cccc}
\hline
\multirow{2}{*}{\textbf{HPQ (\%)}} &\multirow{2}{*}{\textbf{Non-Probing}} &\multicolumn{4}{c}{\textbf{Probing (w.r.t. $\gamma$)}} \\
& &$0.0$ &$0.2$ &$0.5$ &$1.0$ \\
\hline
\thead{SegFormer (B0) \\+ CondInst (R50)} &56.0 &57.0 &56.8 &56.7 &56.2 \\ \hline
\thead{SegFormer (B5) \\+ CondInst (R50)$^\star$} &61.6 &62.7 &62.6 &62.2 &61.7 \\
\hline
\end{tabular}}
\end{table}

We additionally report the HPQ results on the CPP dataset with probing-based inference in Table~\ref{tab:cpp-overall} (while only non-probing-based results are reported in Table~\ref{tab:cpp-comparison}). We have by default used the instance segmentation results with CLIP and mask sampling strategy. Interestingly, although the non-probing-based setting surpasses the probing-based settings with large $\gamma$ values in instance AP, it reports lower HPQ values because probing-based tests usually generate some false positives with low confidence scores -- HPQ, compared to AP, penalizes more on these prediction errors.

\subsection{Details of Competitors}
\label{cpp_competitor}

We introduce the details of the competitors~\cite{degeus2021part,li2022panoptic} compared in Table~\ref{tab:cpp-comparison}, which are the only two existing works that have reported results on CPP. PPS\cite{degeus2021part} first provided the CPP dataset and the PartPQ metric for the task of part-aware panoptic segmentation, and established several baselines by merging results of methods for the subtasks of panoptic and part segmentation. Panoptic-PartFormer~\cite{li2022panoptic} is a unified Transformer-based model that predicts different levels of masks jointly, which achieves significant improvement over PPS\cite{degeus2021part}.

Note that although we provide a comparison in Table~\ref{tab:cpp-comparison}, but this work is essentially different to these methods, and our work is actually not optimized for higher results on the fully-annotated CPP dataset. The most important advantages of \name are the abilities to learn complex hierarchies from incomplete annotations and adapt to new concepts easily, while the competitors~\cite{degeus2021part,li2022panoptic} do not have. The numerical differences (or benefits) in comparison may originate from various aspects (\textit{e.g.}, backbone, training time), and it is difficult to compare with the competitors in a completely fair manner (\textit{e.g.}, keeping the parameters/FLOPs/training epochs in a similar level) since both the methods are highly complicated.

\subsection{Qualitative Results}
\label{cpp:visualization}

\begin{figure*}
\centering
\includegraphics[width=16cm]{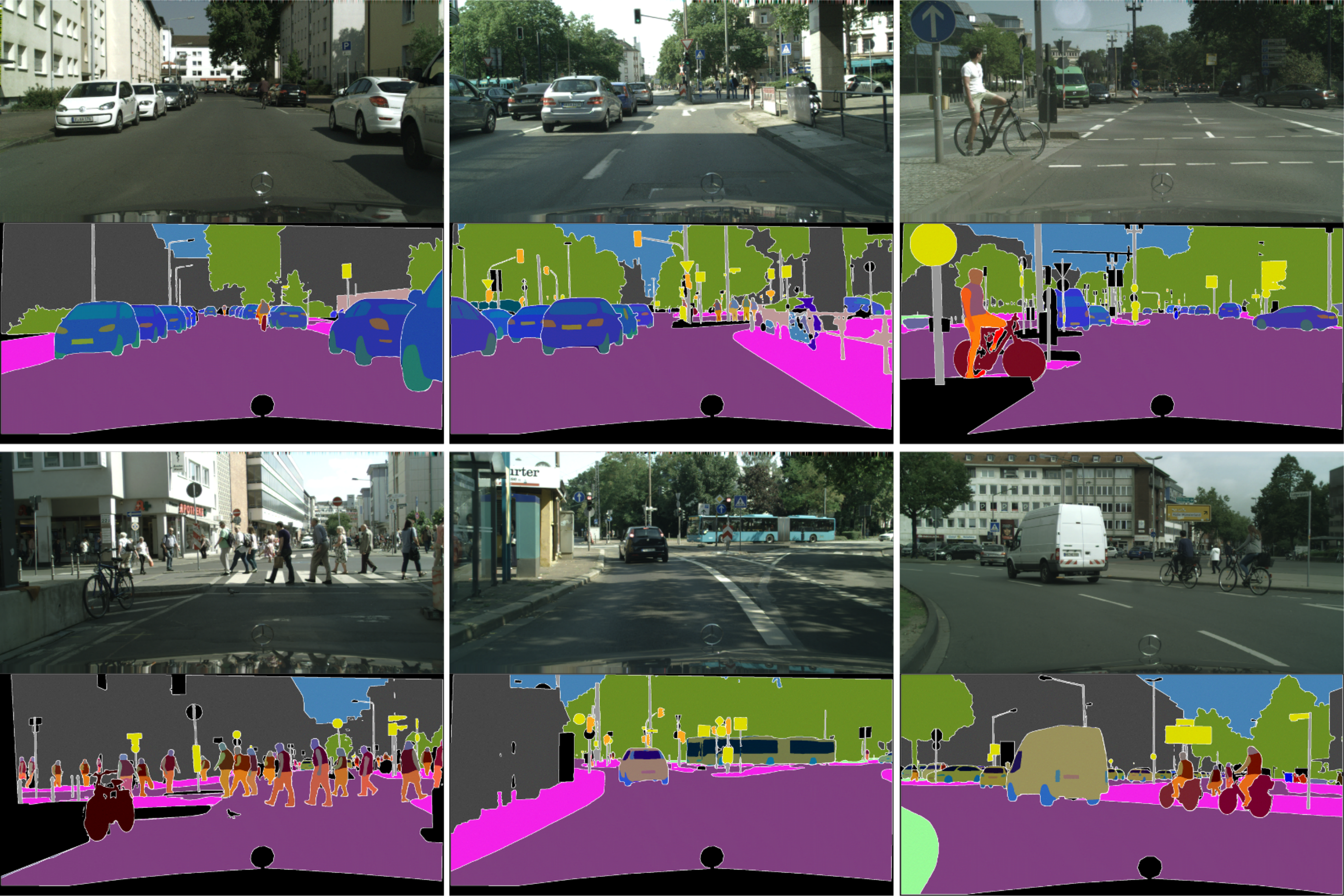}
\vspace{-0.2cm}
\caption{Visualization results of ViRReq on CPP. Results of different requests are merged together. \textit{Best viewed digitally with color.}}
\label{fig:appendix-cpp-vis}
\end{figure*}

We visualize some segmentation results of \name on CPP, as shown in Figure~\ref{fig:appendix-cpp-vis}. The results are obtained through probing-based inference, where the mass center was used as the probe (\textit{i.e.}, $\gamma=0$). The probing pixels are omitted in the figure for simplicity.

\subsection{Open-Domain Recognition}
\label{cpp_open_domain}

\begin{figure*}[!t]
\centering
\includegraphics[width=16cm]{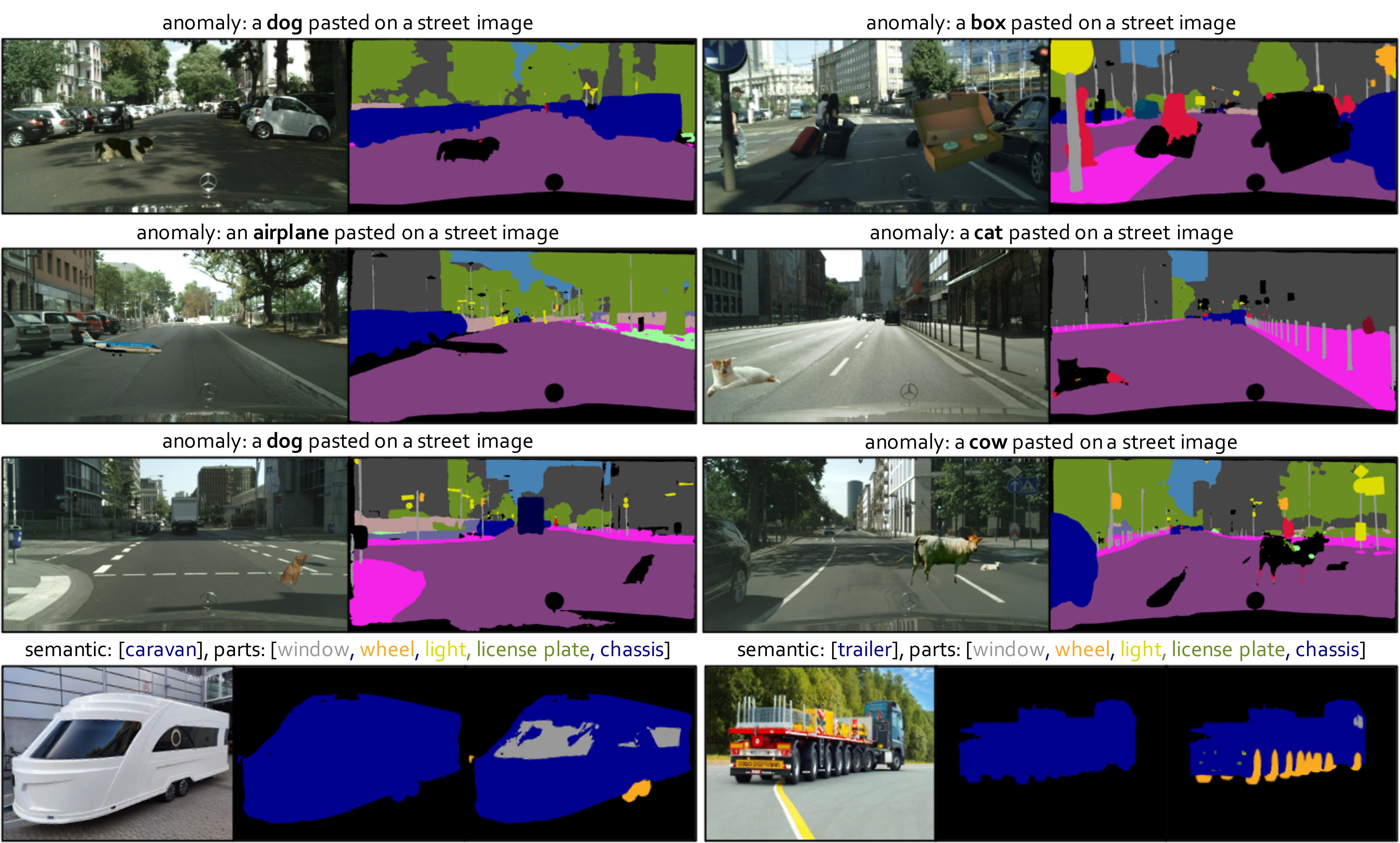}
\vspace{-0.2cm}
\caption{Examples of open-domain recognition on the CPP dataset: anomaly segmentation (top three rows), and compositional segmentation (the last row), respectively. Black region corresponds to \textsf{others}.}
\label{fig:appendix-open-domain}
\end{figure*}

We provide some open-domain segmentation results on CPP, as shown in Figure~\ref{fig:appendix-open-domain}. The top three rows indicate anomaly segmentation, \textit{i.e.}, finding unknown concepts in an image. Example images are taken from the Fishyscapes~\cite{blum2021fishyscapes} dataset (an anomaly segmentation benchmark). These anomalies (\textit{e.g.}, \textsf{dog}, \textsf{box}, \textit{etc.}) were usually not detected by the regular segmentation model (see Figure~2 of \cite{blum2021fishyscapes}), but were found by our approach which was not specifically designed for this task.
The last row involves compositional segmentation, \textit{i.e.}, transferring part-level knowledge from one class (\textit{e.g.}, \textsf{car}) to others (\textit{e.g.}, \textsf{caravan} or \textsf{trailer}) without annotating new data but directly copying the sub-knowledge from the old class to new classes.

\section{Details of the ADE20K Experiments}
\label{ade20k}

\subsection{Data Preparation and Statistics}
\label{ade20k:preparation}

The ADE20K~\cite{zhou2019semantic} dataset provides pixel-wise annotations for more than $3\rm{,}000$ semantic classes, including instance-level and part-level annotations. SceneParse150 is a widely used subset of ADE20K for semantic segmentation, which consists of $20\rm{,}210$ training and $2\rm{,}000$ validation images covering $150$ most frequent classes. For instance segmentation, $100$ foreground object classes are chosen from $150$ classes, termed InstSeg100 (see Section 3.4 of~\cite{zhou2019semantic}), while few works have reported results on InstSeg100. As for part segmentation, we found the annotations are significantly sparse and incomplete~\footnote{The statistics are conducted based on the newest version of ADE20K from the official website.}. There are $289$ non-duplicate part classes belonging to the $100$ instance classes~\footnote{In our definition, only instance classes have part-level annotations.}. The labeling ratio for part classes is pretty low: only $15\%$ of instances have part annotations on averaged ($0.03\% \sim 69.3\%$ for each instance class individually). In our experiments, we first filter the part-of-objects classes that the number of occurrences is no fewer than $100$, resulting in $82$ non-duplicate part-level classes belonging to $40$ instance classes. Then, we further filter the part-of-parts classes that the number of occurrences is no fewer than $100$, resulting in $29$ non-duplicate part-of-parts classes belonging to $17$ upper-level part classes. We conjecture that the sparse annotation property is the main reason that no prior works have reported qualitative results for part-level segmentation on this dataset. Results on the ADE20K validation set are reported. For the vocabulary used in the experiments, semantic and instance classes can be easily found in the original dataset~\cite{zhou2019semantic}, and we additionally list the part-of-object classes as follows in the format of \textit{instance class name (the number of part classes): [part class names]}.

\begin{itemize}
  \setlength\itemsep{0em}
  \small
  \item \textsf{bed (4): [footboard, headboard, leg, side rail]}
  \item \textsf{windowpane (5): [pane, upper sash, lower sash, sash, muntin]}
  \item \textsf{cabinet (7): [drawer, door, side, front, top, skirt, shelf]}
  \item \textsf{person (13): [head, right arm, right hand, left arm, right leg, left leg, right foot, left foot, left hand, neck, gaze, torso, back]}
  \item \textsf{door (5): [door frame, knob, handle, pane, door]}
  \item \textsf{table (4): [drawer, top, leg, apron]}
  \item \textsf{chair (7): [back, seat, leg, arm, stretcher, apron, seat cushion]}
  \item \textsf{car (9): [mirror, door, wheel, headlight, window, license plate, taillight, bumper, windshield]}
  \item \textsf{painting (1): [frame]}
  \item \textsf{sofa (7): [arm, seat cushion, seat base, leg, back pillow, skirt, back]}
  \item \textsf{shelf (1): [shelf]}
  \item \textsf{mirror (1): [frame]}
  \item \textsf{armchair (9): [back, arm, seat, seat cushion, seat base, earmuffs, leg, back pillow, apron]}
  \item \textsf{desk (1): [drawer]}
  \item \textsf{wardrobe (2): [door, drawer]}
  \item \textsf{lamp (9): [canopy, tube, shade, light source, column, base, highlight, arm, cord]}
  \item \textsf{bathtub (1): [faucet]}
  \item \textsf{chest of drawers (1): [drawer]}
  \item \textsf{sink (2): [faucet, tap]}
  \item \textsf{refrigerator (1): [door]}
  \item \textsf{pool table (3): [corner pocket, side pocket, leg]}
  \item \textsf{bookcase (1): [shelf]}
  \item \textsf{coffee table (2): [top, leg]}
  \item \textsf{toilet (3): [cistern, lid, bowl]}
  \item \textsf{stove (3): [stove, oven, button panel]}
  \item \textsf{computer (4): [monitor, keyboard, computer case, mouse]}
  \item \textsf{swivel chair (3): [back, seat, base]}
  \item \textsf{bus (1): [window]}
  \item \textsf{light (5): [shade, light source, highlight, aperture, diffusor]}
  \item \textsf{chandelier (4): [shade, light source, bulb, arm]}
  \item \textsf{airplane (1): [landing gear]}
  \item \textsf{van (1): [wheel]}
  \item \textsf{stool (1): [leg]}
  \item \textsf{microwave (1): [door]}
  \item \textsf{sconce (5): [shade, arm, light source, highlight, backplate]}
  \item \textsf{traffic light (1): [housing]}
  \item \textsf{fan (1): [blade]}
  \item \textsf{monitor (1): [screen]}
  \item \textsf{glass (4): [opening, bowl, base, stem]}
  \item \textsf{clock (1): [face]}
\end{itemize}

The part-of-parts classes are listed as follows in the format of \textit{part class name (the number of part-of-parts classes): [part-of-parts class names]}.

\begin{itemize}
  \setlength\itemsep{0em}
  \small
  \item \textsf{upper sash (3): [pane, stile, muntin]}
  \item \textsf{lower sash (4): [rail, stile, pane, muntin]}
  \item \textsf{sash (4): [pane, rail, stile, muntin]}
  \item \textsf{drawer (2): [knob, handle]}
  \item \textsf{door (7): [hinge, knob, handle, pane, mirror, window, muntin]}
  \item \textsf{head (5): [eye, mouth, nose, ear, hair]}
  \item \textsf{back (3): [rail, spindle, stile]}
  \item \textsf{arm (4): [inside arm, outside arm, arm panel, arm support]}
  \item \textsf{wheel (1): [rim]}
  \item \textsf{window (3): [muntin, pane, shutter]}
  \item \textsf{column (2): [shaft, capital]}
  \item \textsf{base (1): [wheel]}
  \item \textsf{stove (1): [burner]}
  \item \textsf{oven (1): [door]}
  \item \textsf{button panel (1): [dial]}
  \item \textsf{monitor (1): [screen]}
  \item \textsf{face (1): [hand]}
\end{itemize}

\subsection{More Quantitative Results}
\label{ade20k:moer_quantitative}

\begin{table}[!t]
\centering
\caption{Semantic segmentation (Type-I) accuracy of ADE20K on Level-1 (\textit{i.e.}, scene classes, \textit{e.g.}, \textsf{car}) and Level-2 (\textit{i.e.}, part-of-object classes, \textit{e.g.}, \textsf{wheel}) classes, Level-3 (\textit{i.e.}, part-of-parts classes, \textit{e.g.}, \textsf{rim}).}
\vspace{-0.2cm}
\label{tab:ade-individual}
\resizebox{0.8\linewidth}{!}{
\renewcommand{\arraystretch}{1.2}
\setlength{\tabcolsep}{0.08cm}
\begin{tabular}{l|ccc}
\hline
\textbf{mIoU} (\%) & \textbf{Lv-1} & \textbf{Lv-2} & \textbf{Lv-3} \\ \hline
SegFormer (B0)~\cite{xie2021segformer} & \textbf{37.85} & -- & -- \\
w/ CLIP \& parts (ours) & 36.38 & 43.08 & 51.56 \\ \hline
SegFormer (B5)~\cite{xie2021segformer} & \textbf{49.13} & -- & -- \\
w/ CLIP \& parts (ours) & 48.52 & 55.13 & 63.40 \\ \hline
\end{tabular}}
\end{table}

\begin{table}[!t]
\centering
\caption{Instance segmentation results with non-probing-based and probing-based inference on the ADE20K dataset. Results are produced using the semantic mask prediction of SegFormer-B5. The hyper-parameter $\gamma$ controls the centerness of the probes.}
\vspace{-0.2cm}
\label{tab:appendix-mask-sampling-ade}
\setlength{\tabcolsep}{0.12cm}
\renewcommand{\arraystretch}{1.2}
\resizebox{\linewidth}{!}{
\begin{tabular}{l|ccccc}
\hline 
\multirow{2}{*}{\textbf{AP} (\%)} & \multirow{2}{*}{\textbf{Non-Probing}} & \multicolumn{4}{c}{\textbf{Probing (w.r.t. $\gamma$)}} \\
& & $0.0$ & $0.2$ & $0.5$ & $1.0$ \\
\hline 
CondInst (R50)~\cite{tian2020conditional} &24.6 & -- & -- & -- & -- \\
w/ CLIP \& mask samp. & \textbf{25.2} & \textbf{29.8} & \textbf{29.4} & \textbf{27.6} & \textbf{24.0} \\
\hline 
\end{tabular}}
\end{table}

\begin{table}[!t]
\caption{Overall segmentation results on ADE20K, using non-probing-based inference and probing-based inference respectively.}
\vspace{-0.2cm}
\label{tab:ade-overall}
\centering
\setlength{\tabcolsep}{0.12cm}
\renewcommand{\arraystretch}{1.2}
\resizebox{0.9\linewidth}{!}{
\begin{tabular}{l|c|cccc}
\hline
\multirow{2}{*}{\textbf{HPQ (\%)}} &\multirow{2}{*}{\textbf{Non-Probing}} &\multicolumn{4}{c}{\textbf{Probing (w.r.t. $\gamma$)}} \\
& &$0.0$ &$0.2$ &$0.5$ &$1.0$ \\
\hline
\thead{SegFormer (B0) \\+ CondInst (R50)} &27.2 &34.2 &34.2 &33.8 &32.9 \\ \hline
\thead{SegFormer (B5) \\+ CondInst (R50)} &33.9 &39.1 &38.9 &38.5 &37.9 \\
\hline
\end{tabular}}
\end{table}

We report the individual segmentation accuracy of Type-I and Type-II requests on ADE20K, as shown in Tables~\ref{tab:ade-individual} and~\ref{tab:appendix-mask-sampling-ade}, respectively.
We additionally report the HPQ results on the ADE20K dataset with probing-based inference in Table~\ref{tab:ade-overall} (while only non-probing-based results are reported in Table~\ref{tab:ade-comparison}). We have by default used the instance segmentation results with CLIP and mask sampling strategy. Similar to the observations on CPP, although the non-probing-based setting achieves comparable instance AP, it reports lower HPQ values due to the penalty on false positives. Besides, all these values are significantly lower than the values reported on CPP, indicating that ADE20K is much more challenging in terms of the richness of semantics and the granularity of annotation. Overall, there is much room for improvement in such a complex, multi-level, and sparsely annotated segmentation dataset.

\subsection{Details of Competitor}
\label{ade_competitor}

\begin{figure}
\centering
\includegraphics[width=\linewidth]{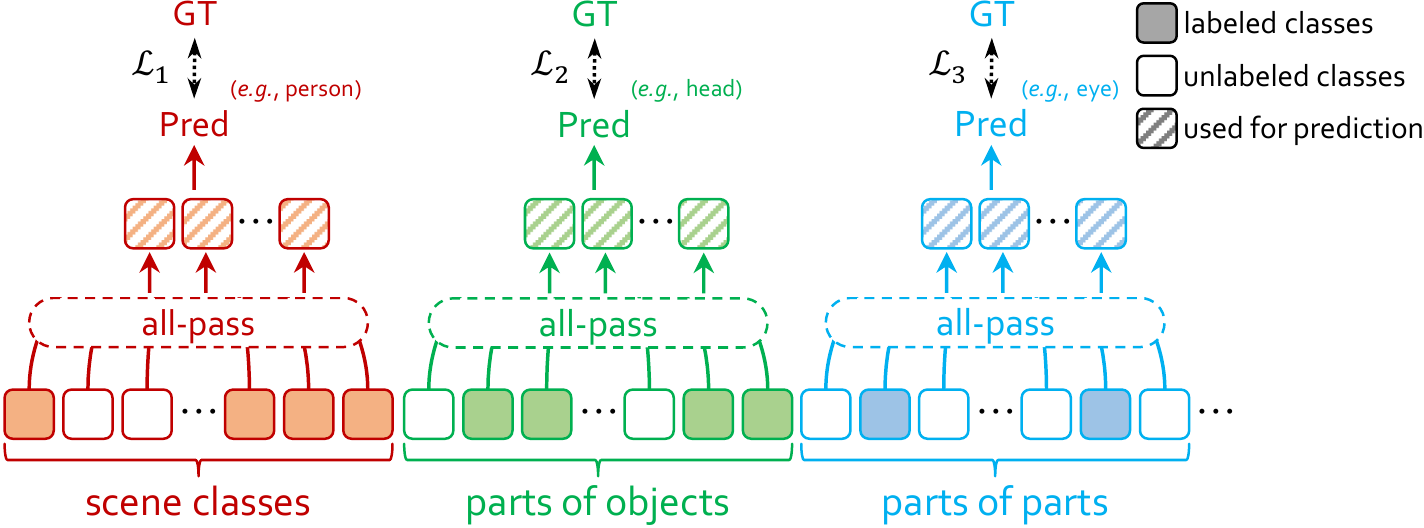}
\vspace{-0.6cm}
\caption{An illustration of how Type-I requests are processed for \textit{one pixel} with the \textbf{conventional method}. Each slot indicates the predicted logit of one class. All classes are used during training. See Figure~\ref{fig:group_loss} for detailed differences.}
\label{fig:appendix-ade20k-competitor}
\end{figure}

We introduce the details of the competitor (\textit{i.e}, conventional method) compared in Table~\ref{tab:ade-comparison}. Since no prior works have ever reported quantitative results for part-aware segmentation on ADE20K, we provide a preliminary solution of adapting conventional methods to ADE20K (see Section~\ref{experiments:ade}). Figure~\ref{fig:appendix-ade20k-competitor} illustrates how the competitor method deals with Type-I requests, \textit{i.e.}, joint training in a multi-task manner and simply ignoring all unlabeled pixels. There are several main differences compared to \name: (i) using fixed class ID and learnable classifier for each class instead of language-based queries, (ii) predicting different levels of results simultaneously instead of iteratively, and (iii) how to handle unlabeled pixels. See Figure~\ref{fig:group_loss} for detailed differences.

\subsection{More Qualitative Results}
\label{ade20k:more_qualitative}

\begin{figure*}[!t]
\centering
\includegraphics[width=\linewidth]{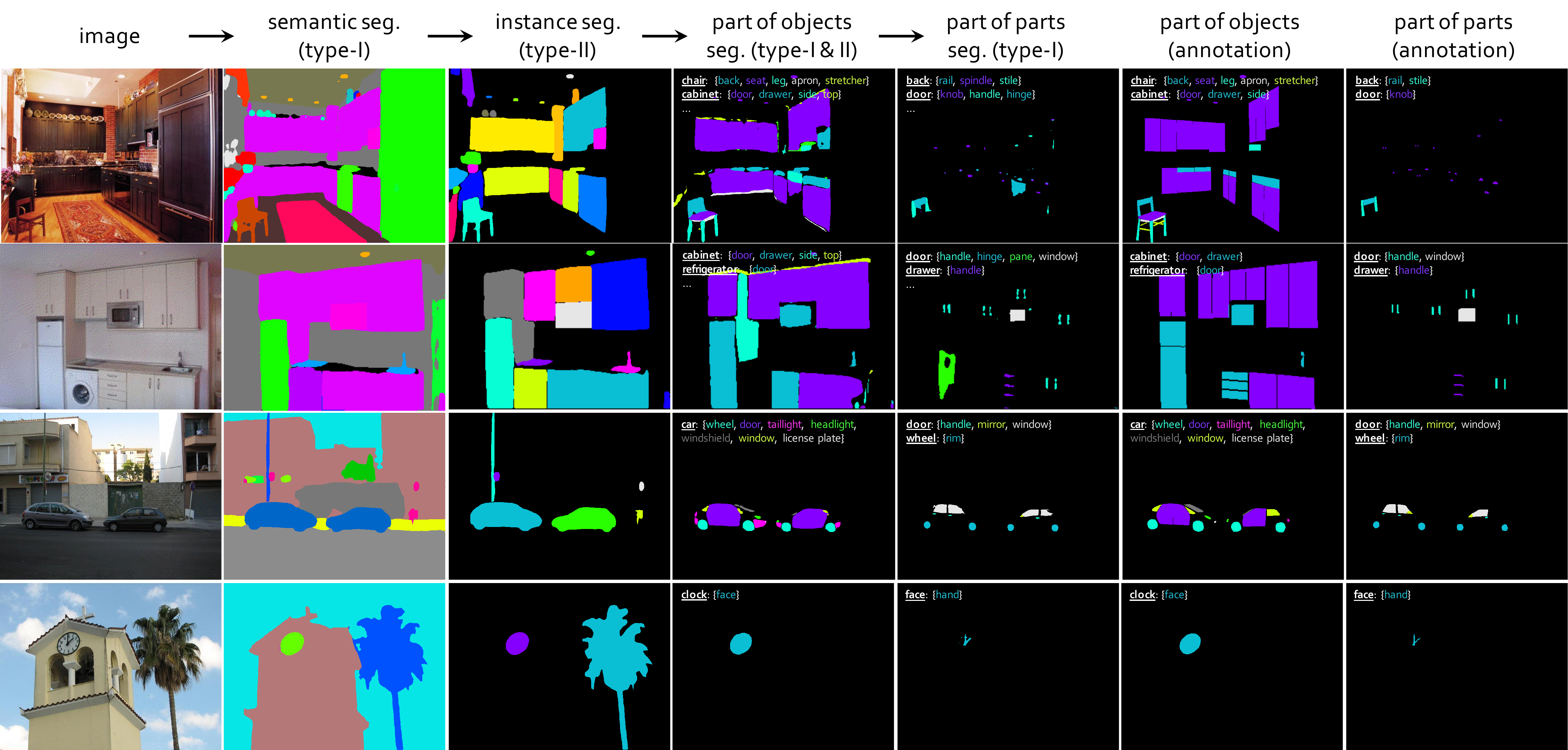}
\vspace{-0.6cm}
\caption{More visualization results of \name on ADE20K. Black areas in prediction indicate the \textsf{others} (\textit{i.e.}, unknown) class. The corresponding sub-knowledge is listed in the blank area for reference. \textit{Best viewed in color and by zooming in for details.}}
\label{fig:appendix-ade-vis}
\end{figure*}

We visualize more segmentation results of \name on ADE20K, as shown in Figure~\ref{fig:appendix-ade-vis}.

\section{Details of Few-shot Incremental Learning}
\label{more_on_incremental}

\name has the ability of learning new visual concepts (objects and/or parts) from a few training samples. To show this ability, we manually select $50$ scene-level and $19$ part-level long-tailed classes of ADE20K and add them to the original knowledge base, as listed below. For each new concept, $20$ instances are labeled for few-shot learning.

\begin{itemize}
  \setlength\itemsep{0em}
  \small
  \item \textsf{\textbf{scene-level classes:} spotlight, wall socket, fluorescent, jar, shoe, candlestick, switch, air conditioner, telephone, mug, container, board, candle, cup, pitcher, deck chair, light bulb, coffee maker, teapot, partition, shrub, figurine, magazine, can, umbrella, bucket, napkin, text, gravestone, pane, patty, manhole, hat, doorframe, curb, loudspeaker, snow, pool ball, hedge, pipe, central reservation, booklet, grill, place mat, faucet, notebook, document, fish, jacket, price tag}
  \item \textsf{\textbf{part-level classes:} bedpost, sill, casing, rail, stile, gas cap, chain, capital, shaft, bed, burner, dial, speaker, piston, wing, stabilizer, fuselage, turbine engine, motor}
\end{itemize}

\begin{figure}[!t]
\centering
\includegraphics[width=\linewidth]{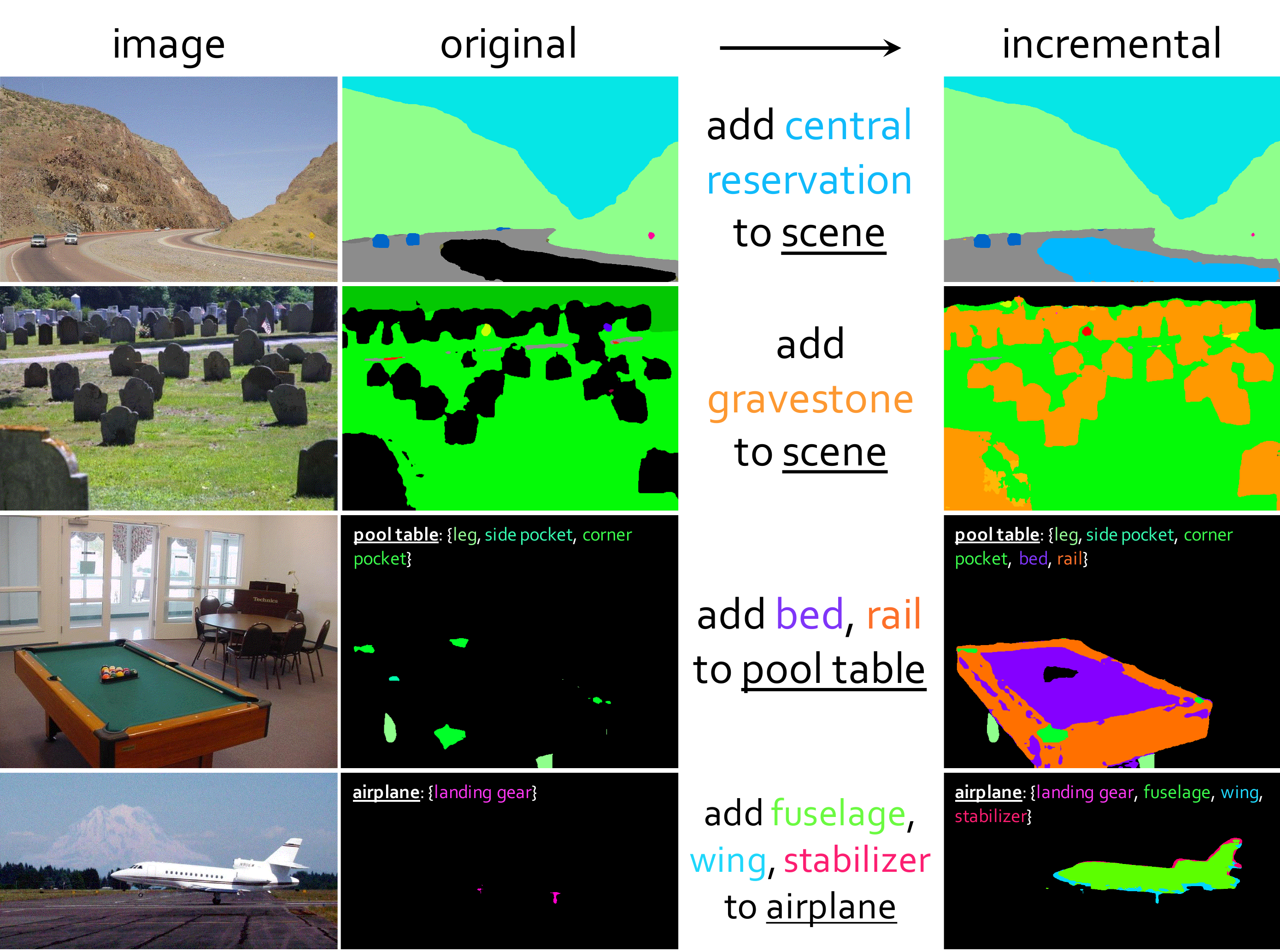}
\vspace{-0.6cm}
\caption{More qualitative results of few-shot incremental learning on object parts (top two rows), and parts of parts (bottom two rows). Prior to incremental learning, these new concepts are recognized as \textsf{others} (the black areas) which is as expected.}
\label{fig:appendix-ade-incremental}
\end{figure}

We noticed that existing few-shot semantic segmentation approaches~\cite{Min2021HypercorrelationSF,Dong2018FewShotSS,Wang2019PANetFI,Liu2020PartawarePN} usually focus on finding visual correspondences between query and support images for mask prediction (\textit{e.g.,} the prototype-based methods), without updating the original models learned on base classes. However, we believe that it is more essential to update (\textit{e.g.}, fine-tune) the model to learn knowledge of the new visual concepts. Inspired by the recent works on CLIP-based few-shot classification~\cite{gao2021clip,zhou2022learning}, we mostly follow CLIP-Adapter~\cite{gao2021clip} to integrate an additional bottleneck layer (\textit{e.g.}, two $1\times1$ convolutional layers) for feature alignment. Only this bottleneck layer gets updated during the fine-tuning, \textit{i.e.}, learning to project the original trained visual features to a new feature space.

Note that, in our setting, the existing training images may contain unlabeled pixels of these new classes (expect to recognize as \textsf{others} originally). To avoid possible conflict, we associate each training image to the knowledge base that was used for annotation (which we call data versioning, see Section~\ref{problem:advantages}). For example, one image may have multiple copies in the training set that associated with different version of knowledge base (\textit{e.g.}, annotating for the newly added concepts or not).

Specifically, we mix the original and new training images and fine-tune the trained SegFormer-B5 model for $40\mathrm{K}$ iterations ($1/4$ of base training iterations). Since the number of new training images (\textit{e.g.}, $20 \times 50 = 1000$ images for scene-level incremental learning) is significantly lower than the original (\textit{e.g.}, $20\rm{,}210$ training images), these new images are repeated multiple times during training to alleviate the imbalance problem. Other configurations are the same as the base training (see Appendix~\ref{details:optimization}). We show more qualitative results in Figure~\ref{fig:appendix-ade-incremental}.


\end{document}